%% file: main-arxiv.tex
\documentclass[10pt,twocolumn,letterpaper]{article}

\usepackage[pagenumbers]{cvpr}

\input{common/preamble}
\definecolor{cvprblue}{rgb}{0.21,0.49,0.74}
\usepackage[breaklinks,colorlinks,allcolors=cvprblue]{hyperref}
\def\confName{arXiv}
\def\confYear{2027}

\input{common/macros}
\renewcommand{\releasewhen}{}
\renewcommand{\releasenote}{Released with this paper.}

\renewcommand{\aftercap}[1]{\vspace{-1.9mm}}
\usepackage{fontawesome5}
\usepackage{twemojis}

\newcommand{\linkpage}{wongyun-yu.github.io/dexengine}
\newcommand{\linkcode}{github.com/wongyun-yu/DexEngine}
\newcommand{\linkdata}{huggingface.co/datasets/Wongyun/DexEngine}
\newcommand{\weblink}[2]{#1\,\href{https://#2}{#2}}

\input{common/title}
\makeatletter\def\@title{DexEngine: Scaling Human Demonstrations\texorpdfstring{\\}{ }into Physically Executable Whole-Robot Data}\renewcommand{\thetitle}{DexEngine: Scaling Human Demonstrations\texorpdfstring{\\}{ }into Physically Executable Whole-Robot Data}\makeatother
\author{%
Wongyun Yu$^{1,2}$\quad Honggeun Kim$^{1}$\quad Donguk Lee$^{2}$\\
Minchan Kim$^{2}$\quad Kyoung Whan Choe$^{2}$\quad Minsu Cho$^{1,2}$$^{\dagger}$\\[2pt]
$^{1}$POSTECH\qquad $^{2}$RLWRLD
}
\newcommand{\titlelinks}{\weblink{\faGlobe}{\linkpage}\hspace{1.5em}\weblink{\faGithub}{\linkcode}\hspace{1.5em}\weblink{\texttwemoji{1f917}}{\linkdata}}

\makeatletter
\def\@maketitle{\newpage\null\kern -8pt
  \begin{center}{\Large \bf \@title \par}\vspace*{16pt}{\large\lineskip .5em\begin{tabular}[t]{c}\@author\end{tabular}\par}
  \vspace*{4.07pt}{\small\titlelinks\par}\vspace*{5.24pt}\end{center}}
\makeatother
\begin{document}
\maketitle
{\renewcommand{\thefootnote}{\textdagger}\footnotetext{Corresponding author.}}
\input{common/body}

\long\def\acktext{%
We would like to thank Se June Joo of RLWRLD for insightful discussions and thoughtful support.
}
\if\relax\detokenize\expandafter{\acktext}\relax\else\section*{Acknowledgments}\acktext\par\fi
{\small
\bibliographystyle{ieeenat_fullname}
\bibliography{ref}
}
\twocolumn[{\centering{\Large\bfseries Supplementary Material}\par\vspace{1.5em}}]
\input{common/supp}

\end{document}

%% file: common/preamble.tex
\usepackage{cuted}
\usepackage{bm}
\usepackage{placeins}
\usepackage{array}
\usepackage{tikz}
\DeclareMathSizes{8}{8}{7}{7}
\DeclareMathSizes{9}{9}{7}{7}


\graphicspath{{figs_tabs/figures/}{assets/}}

%% file: common/macros.tex
\newcommand{\releasewhen}{ upon publication}
\newcommand{\aftercap}[1]{\vspace{#1}}
\newcommand{\releasenote}{Upon publication.}

\newcommand{\fitwidth}[1]{\sbox0{#1}\ifdim\wd0>\linewidth\resizebox{\linewidth}{!}{\usebox0}\else\usebox0\fi}
\newcommand{\chip}[1]{\textcolor[HTML]{#1}{\rule{4pt}{4pt}}}

\newlength{\panelw}

\newcommand{\mkfloat}{{\definecolor{mk}{HTML}{EB6834}\tikz[baseline=-0.6ex]{\fill[mk] (0,0) circle (2pt);}}}
\newcommand{\mkarm}{{\definecolor{mk}{HTML}{2A78D6}\definecolor{mkl}{HTML}{8AB5E8}\tikz[baseline=-0.6ex]{\draw[line width=0.6pt,draw=mk,fill=mkl] (0,0) circle (1.9pt);}}}
\newcommand{\mkarmk}{{\definecolor{mk}{HTML}{2A78D6}\tikz[baseline=-0.6ex]{\fill[mk] (0,0) circle (2pt);}}}
\definecolor{dcReal}{HTML}{CDEBC5}\definecolor{dcSim}{HTML}{CFE0F5}\definecolor{dcNone}{HTML}{F3D3CF}
\definecolor{dcBi}{HTML}{CDEBC5}\definecolor{dcUni}{HTML}{FBE7BD}\definecolor{dcHand}{HTML}{E4E4E4}
\newcommand{\dcbox}[3]{{\setlength{\fboxsep}{1.2pt}\colorbox{#1}{\makebox[#2][c]{\strut #3}}}}
\newcommand{\dcR}{\dcbox{dcReal}{1.9em}{Real}}\newcommand{\dcS}{\dcbox{dcSim}{1.9em}{Sim}}\newcommand{\dcN}{\dcbox{dcNone}{1.9em}{None}}
\newcommand{\dcB}{\dcbox{dcBi}{1.9em}{Bi.}}\newcommand{\dcU}{\dcbox{dcUni}{1.9em}{Uni.}}\newcommand{\dcH}{\dcbox{dcHand}{1.9em}{Hand}}

%% file: common/title.tex
\title{DexEngine: Scaling Human Demonstrations into Physically Executable Whole-Robot Data}

%% file: common/body.tex
\input{sections/00_abstract}
\input{sections/10_intro}

\input{sections/11_contributions}
\input{sections/20_related}
\input{sections/30_method_overview}

\input{sections/31_method_preparation}

\input{sections/32_method_reference}

\input{sections/33_method_rl}

\input{sections/34_method_data}

\input{sections/40_exp_dataset}

\input{sections/41_exp_pipeline}
\input{sections/42_exp_policy}
\input{sections/43_exp_real}
\input{sections/50_conclusion}

\FloatBarrier

%% file: sections/00_abstract.tex
\begin{abstract}
{\looseness=0
A growing body of recent work addresses the retargeting of human hand–object demonstrations to robotic hands. To train whole-robot policies, however, the retargeted motions must be kinematically feasible and physically executable, and be produced efficiently at scale.
We present DexEngine, a simulation pipeline that turns human manipulation data into train-ready, physically executable arm–hand trajectories. DexEngine first places each clip’s hand–object motion where the arms can reach it, finds an arm motion that follows the human hands, and then corrects the motion in physics simulation with residual reinforcement learning. From nine human hand–object datasets, we generate about 5,200 clips for each of four dexterous hands, augmented into 65,000 trajectories (116 h) and 325,000 episodes (578 h) in total. Even with arms attached, DexEngine tracks objects more accurately than prior floating-hand methods and runs over 10$\times$ faster than RL-based ones. Policies trained on our data far outperform those trained on the same clips without physics correction or with the arms attached after a floating-hand correction, showing that fine-tuning data need physics correction. Mid-training on our data further improves them at equal fine-tuning budget, in simulation \mbox{and on a real robot}.
\par}
\end{abstract}

%% file: sections/10_intro.tex
\section{Introduction}
\label{sec:intro}
Dexterous robot policies learn from demonstrations, which are slow and costly to collect by teleoperation~\cite{mimicgen,dexmimicgen,unidex}.
Human hand--object interaction, in contrast, is easier to capture~\cite{hot3d,oakink2,taco,grab,arctic,h2o,hocap,dexcanvas} and covers more diverse tasks than teleoperation~\cite{huro,egoscale}.
Many recent methods therefore convert human data into robot data~\cite{unidex,huro,egoengine,maniptrans,dexmachina,spider,chord,flashdexretarget,oath,dexman}.
Kinematic methods map human hand keypoints to robot joint angles by optimization or inverse kinematics (IK)~\cite{toporetarget,unidex}. Physics-based methods search for a robot motion that reproduces the object motion in simulation, using sampling-based optimization~\cite{spider} or reinforcement learning (RL)~\cite{maniptrans,dexmachina,chord}.
Several of these methods build robot datasets this way~\cite{unidex,huro,spider,maniptrans}, and UniDex~\cite{unidex} and HuRo~\cite{huro} pretrain policies on them.
\input{sections/floats/fig_teaser}
\input{sections/floats/fig_pipeline}
Yet these data fall short as training data for whole-robot policies in three ways.
First, kinematic methods~\cite{unidex,huro} ignore dynamics: without contact and force, the robot may fail to move the object, and HuRo~\cite{huro} uses its trajectories as pretraining supervision only.
Second, most physics-based methods~\cite{maniptrans,dexmachina,chord,spider,dextrack,flashdexretarget} ignore the arm: they drive floating hands and leave the arm's joint limits, reach, and collisions unchecked, even though a policy must control both arm and hand joints.
Third, physics-based methods do not scale: the few that include the arm~\cite{oath,dexman,gnr} cover fewer than a thousand human demonstrations, and most are either slow or inaccurate on many clips.

We present \textbf{DexEngine}, a scalable pipeline that turns human capture data into physically executable, train-ready simulation data for bimanual robots with dexterous hands.
It rests on two key ideas.
First, it solves the arms before physics: it searches each clip's scene placements for one that the arms can follow, and keeps nearly every clip that has such a placement.
By contrast, attaching the arms after a floating-hand correction keeps 26\% fewer clips, even at our placements (\cref{sec:exp_data}).
Second, it reuses this arm IK solution in residual RL, whose policy acts on the wrist pose in task space and never observes the arm joints.
Since IK has multiple solutions for a wrist pose, each random start~\cite{deepmimic} resets the arm to this solution instead of re-solving IK, which would switch branches and lower accuracy (\cref{tab:abl_rl}).
With this recipe, even with the arms attached, DexEngine corrects up to 64 clips with one policy, more accurately than existing floating-hand methods, and is the fastest training-based method, at 3.9 GPU-minutes per clip (\cref{tab:same64}).

Our contributions are fourfold:

%% file: sections/floats/fig_teaser.tex
\begin{figure}[t]
  \centering
  \includegraphics[width=\linewidth]{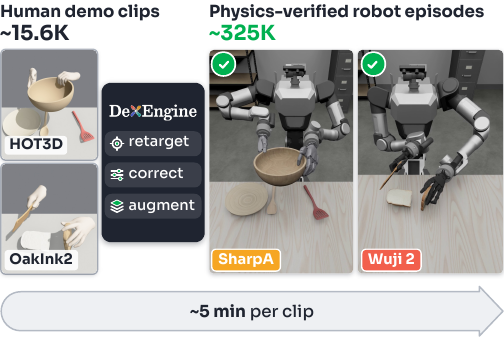}
  \caption{\textbf{DexEngine} turns human demonstrations into physically executable data for a whole bimanual robot, in under 5 GPU-minutes per clip.}   
  \aftercap{-4.6mm}
  \label{fig:teaser}
\end{figure}

%% file: sections/floats/fig_pipeline.tex
\begin{figure*}[t]
  \centering
  \includegraphics[width=\linewidth]{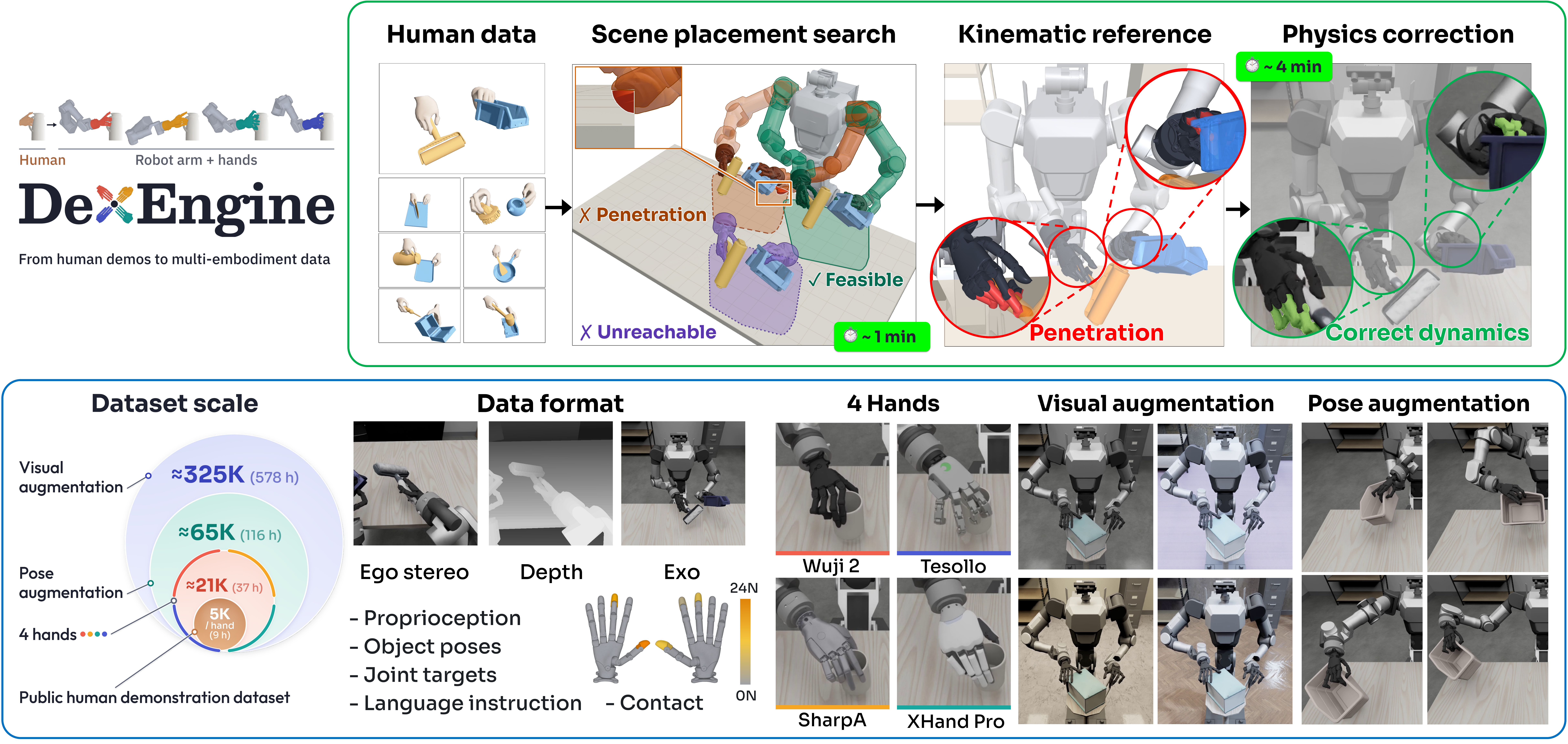}
  \caption{\textbf{DexEngine pipeline.} Human demonstrations are split into clips, a scene placement that the arms reach without penetration is found, and the resulting kinematic reference is corrected in physics simulation. The corrected motions become train-ready data for four robot hands, with pose and visual augmentation. Red marks penetration in the kinematic reference, and green the corrected contact.}
  \aftercap{-1.9mm}
  \label{fig:pipeline}
\end{figure*}

%% file: sections/11_contributions.tex
\begin{enumerate}
  \item DexEngine, a scalable pipeline that turns human demonstrations into simulation data for a whole bimanual robot by solving the arm motion before physics (\cref{sec:method}).
  \item An arm attachment for residual RL, acting in wrist space~\cite{dexman} with the arm reset to the kinematic reference at random starts, that keeps the policy accurate and lets it run on a rigidly transformed scene without retraining (\cref{tab:abl_rl,sec:data}).
  \item Experiments showing that fine-tuning data need physics correction and mid-training on our data helps at equal fine-tuning budget, also on a real robot (\cref{sec:experiments}).
  \item A train-ready, physically executable dataset of 65,000 trajectories (116 h) and 325,000 episodes (578 h) with contact forces and depth for a bimanual robot with each of four dexterous hands, released\releasewhen~(\cref{tab:datacmp}).
\end{enumerate}

%% file: sections/20_related.tex
\section{Related work}
\label{sec:relwork}

\noindent\textbf{Sources of robot data.}
Robot training data are teleoperated, synthesized, or converted from human capture.
Data generators adapt a few teleoperated demonstrations to new scenes~\cite{mimicgen,dexmimicgen}.
Synthesis without human data reaches about one billion demonstrations, but only for grasping and opening articulated objects~\cite{dex1b}.
Human hand--object datasets record hand and object trajectories across diverse tasks, without robot actions~\cite{grab,h2o,arctic,taco,oakink2,hocap,hot3d,dexcanvas}, and HRDexDB~\cite{hrdexdb} adds robot grasps only.
Pipelines from human videos reconstruct such trajectories~\cite{deximit} and then retarget them~\cite{doasido,dexman,egoengine,egocpt,egoinfinity,videomanip}, and RoboPaint~\cite{robopaint} renders robots retargeted from glove captures into reconstructed scenes.
We convert nine human datasets into physically executable arm--hand trajectories.

\noindent\textbf{Human-to-robot retargeting.}
Retargeting methods differ in whether the robot moves the object as demonstrated.
Kinematic retargeting by optimization~\cite{toporetarget} or IK~\cite{unidex} lets UniDex~\cite{unidex} and HuRo~\cite{huro} build large robot datasets from human videos, but without contact forces, the robot may fail to move the object.
RoboWheel~\cite{robowheel} corrects the human hand--object poses in physics before retargeting, but not the retargeted robot motion.
Physics-based retargeting instead finds a robot motion that reproduces the object motion, by sampling-based optimization~\cite{spider} or RL~\cite{maniptrans,dexmachina,chord,egocpt}.
To avoid solving each demonstration separately, recent methods fit one model to many~\cite{flashdexretarget,gnr,oath,contactanchor} or train one to generalize to unseen motions~\cite{dextrack}.
We correct the retargeted motions with one residual policy per bundle of clips and keep only clips whose objects follow the demonstration.

\noindent\textbf{Arms and scene placement.}
An arm-mounted hand can follow a demonstration only if its arm can reach every wrist pose without exceeding joint limits or colliding.
Floating-hand methods~\cite{maniptrans,dexmachina,dextrack} and SPIDER's released data~\cite{spider} ignore these constraints.
EgoEngine~\cite{egoengine} mounts the hand on an arm but refines its motion with a floating wrist.
OATH~\cite{oath}, DexMan~\cite{dexman}, and GNR~\cite{gnr} move the arm to follow the retargeted wrist and refine the motion in simulation, but do not search the scene placement, on which IK success depends.
Prior pipelines set this placement manually~\cite{doasido}, align the robot base kinematically to human videos~\cite{huro}, or pick the candidate base pose with the best IK result~\cite{egoinfinity}.
SPIDER~\cite{spider} samples random placements for a single clip and keeps those where IK follows the whole motion.
We instead search each clip's placements with a reachability map~\cite{capabilitymap}, as used to place robot bases~\cite{reachinversion,reuleaux}, and run IK only on promising ones before simulation.

\noindent\textbf{Policy training on generated and human data.}
Human and robot data enter three stages of policy training: pretraining, mid-training, and task fine-tuning.
UniDex~\cite{unidex}, HuRo~\cite{huro}, and PointWAM~\cite{pointwam} pretrain on human videos, with or without kinematically retargeted actions, and report gains on fine-tuning.
Mid-training adapts the pretrained model to the target robot with robot-matched data~\cite{egoscale,trex,rldx1}.
EgoScale~\cite{egoscale} and T-Rex~\cite{trex} report gains from mid-training with teleoperation data, and RLDX-1~\cite{rldx1} adds synthetic episodes from a video generative model.
In task fine-tuning, most physics-based retargeting methods do not train policies on their output~\cite{dexmachina,chord,flashdexretarget,dexman,oath} or use it for one to four tasks~\cite{spider,maniptrans,egoengine}.
EgoEngine~\cite{egoengine} finds that executable actions improve policies, but tests this only on a single arm.
Other methods train RL policies on human data in simulation and run them on a real robot directly or via a distilled visual policy~\cite{human2sim2robot,xsim,vividex,regrind,dexx}.
We correct whole-robot motions, arms included, across nine human datasets and use the data for both mid-training and task fine-tuning, in simulation and on a real bimanual robot.

%% file: sections/30_method_overview.tex
\section{Method}
\label{sec:method}
\subsection{Overview}
\noindent\textbf{Problem formulation.}
We consider a bimanual robot with a dexterous hand on each arm.
The input is a human hand--object demonstration, a single-task \emph{clip} (\cref{sec:parsing}) that gives at every frame the wrist pose and finger articulation (\eg, MANO~\cite{mano}) of each human hand and the pose of each object, whose mesh is known.
The output is a robot motion $\mathcal{Q}=\{\mathbf{q}_{t}\}$ that meets two requirements, where $\mathbf{q}_{t}$ holds the joints of both arms and hands.
It must be \emph{kinematically feasible}, staying within the joint position and velocity limits with no arm penetrating the support surface, a fixed object, the robot body or the other arm.
It must also be \emph{physically executable}, moving the objects along their demonstrated poses in physics simulation.
Whether a feasible motion exists depends on where the demonstration sits in the arms' limited workspace.
We therefore also choose a \emph{placement} $\boldsymbol{\xi}$, a rigid body transformation of the demonstrated hands and objects relative to the robot base (\cref{sec:reference}).
Let $\mathcal{F}(\boldsymbol{\xi})$ contain the kinematically feasible motions at $\boldsymbol{\xi}$, and let the target pose $\hat{\mathbf{O}}^{j}_{t}\in SE(3)$ of object $j$ at frame $t$ be its demonstrated pose moved by $\boldsymbol{\xi}$ and retimed with the robot motion.
We cast kinematic feasibility as a constraint and physical executability as an objective,
\begin{equation}
\begin{aligned}
\min_{\mathcal{Q},\,\boldsymbol{\xi}}\;\;
& E=\sum_{t}\sum_{j}d\big(\hat{\mathbf{O}}^{j}_{t},\mathbf{O}^{j}_{t}\big)\\
\text{s.t.}\;\;
& \mathcal{Q}\in\mathcal{F}(\boldsymbol{\xi}),
\end{aligned}
\label{eq:goal}
\end{equation}
where $\mathbf{O}^{j}_{t}$ is the simulated object pose when the robot moves along $\mathcal{Q}$ and $d$ is a distance between poses.
From each solution, we then record \emph{train-ready} data, with the robot's own observations and actions for all joints.

\input{sections/floats/tab_datacmp}\noindent\textbf{Pipeline overview.}
\Cref{eq:goal} cannot be solved directly, since its objective needs costly contact simulation.
Reducing the object error first, by correcting floating hands in physics, can drive the wrists to poses that the arms cannot follow, violating the constraint (\cref{sec:exp_data}).
DexEngine therefore satisfies the constraint before physics and then reduces the object error in simulation, in four stages (\cref{fig:pipeline}).
(1)~Data preparation (\cref{sec:parsing}) divides each sequence into clips.
(2)~Kinematic reference generation (\cref{sec:reference}) finds a placement and an arm motion that satisfy the constraint, removing a clip without one.
(3)~Physics correction (\cref{sec:rl}) reduces the object error with residual RL, resetting the arms to the reference at random starts.
(4)~Train-ready data generation (\cref{sec:data}) records each corrected motion with its actions and contact forces and renders it from the robot's stereo head camera and an exocentric camera.
\Cref{fig:funnel} shows the clips that remain after each stage.

%% file: sections/floats/tab_datacmp.tex
\begin{table}[t]
\centering
{\scriptsize\setlength{\tabcolsep}{0.8pt}
\begin{tabular*}{\linewidth}{@{\extracolsep{\fill}}lccccccccr@{}}
\toprule
Data & Physics & Body & Hands & Views & Depth & Contact & Open & Hours (ep.) \\
\midrule
HRDexDB~\cite{hrdexdb} & \dcR & \dcU & 3 & 2+21 & $\times$ & \checkmark & \checkmark & $\approx$11 (1.7K) \\
EgoEngine~\cite{egoengine} & \dcS & \dcH & 1 & 1+0 & $\times$ & $\times$ & $\times$ & n/r \\
UniDex~\cite{unidex} & \dcN & \dcH & 8 & 1+0 & \checkmark & $\times$ & $\times$ & $\approx$83 (52K) \\
HuRo~\cite{huro} & \dcN & \dcB & 1 & 1+0 & $\times$ & $\times$ & \checkmark & 1,317 (630K) \\
SPIDER~\cite{spider} & \dcS & \dcH & 4 & 0 & $\times$ & $\times$ & \checkmark & $\approx$4.3 (7.9K) \\
DexManipNet~\cite{maniptrans} & \dcS & \dcH & 1 & 0 & $\times$ & \checkmark & \checkmark & $\approx$6.2 (3.3K) \\
Do as I Do~\cite{doasido} & \dcS & \dcH & 1 & 0 & $\times$ & $\times$ & $\times$ & n/r (500) \\
GNR~\cite{gnr} (HOT3D) & \dcS & \dcH & 1 & 0 & $\times$ & \checkmark & $\times$ & $\approx$133 (223K) \\
GNR~\cite{gnr} (arm) & \dcS & \dcU & 1 & 0 & $\times$ & \checkmark & $\times$ & $\approx$12 (3.4K) \\
\midrule
DexEngine (Ours) & \dcS & \dcB & 4 & 2+1$^\dagger$ & \checkmark & \checkmark & \checkmark$^\ddagger$ & $\approx$116 (65K) \\
\bottomrule
\end{tabular*}}
\caption{Robot data of human-to-robot pipelines. Physics: Real = real robot, Sim = simulation, None = kinematic only. Body: Hand = floating hand, Uni./Bi. = one/two arms with hands. Views: ego + exo. Contact: contact forces or tactile signals. Open: robot trajectories public. n/r: not reported. $^\dagger$Five appearances per trajectory. $^\ddagger$\releasenote}
\aftercap{-2mm}%
\label{tab:datacmp}
\end{table}

%% file: sections/31_method_preparation.tex
\subsection{Data preparation}
\label{sec:parsing}
DexEngine accepts any sequence of human hand and object poses.
Data preparation thus aligns each sequence to a common frame, divides it into clips with at most one manipulated object per hand, and removes unusable clips. 

\noindent\textbf{Alignment.}
We express each sequence in a common gravity-aligned frame and, where the objects rest on a horizontal support surface such as a table, take its plane from the sequence or set it at the height where they rest (\cref{sec:supp_layout}).

\noindent\textbf{Clip division.}
To reward each hand for its own object in physics correction (\cref{sec:rl}), each hand keeps one manipulated object per clip, the one it moves or whose joint it drives longest (\cref{sec:supp_assignment}).
We therefore cut each sequence where a hand switches objects, except at handovers, and cut long sequences where objects are static (\cref{sec:supp_division}).
Only hands that manipulate an object are active, and the robot holds the arm of an inactive hand, such as the idle hand of a one-handed task, at a rest pose at the side.
The division uses only the human poses, before any retargeting, so that all four robot hands share the same source clips.

\noindent\textbf{Clip filtering.}
We remove clips whose source data are physically inconsistent, such as an object that jumps between frames or sinks into its support surface.
After retargeting (\cref{sec:reference}), we also remove the clips in which a robot hand's fingertips stray from the human ones (\cref{fig:funnel,sec:supp_filtering}).

%% file: sections/32_method_reference.tex
\subsection{Kinematic reference generation}
\label{sec:reference}
For each clip, we compute a \emph{kinematic reference}, a robot motion in $\mathcal{F}(\boldsymbol{\xi})$ that follows the human hands.
We first retarget the human hands to floating robot hands. We then search for a placement at which the arms can follow the wrist poses.
Finally, we retime the arm motion at that placement within the joint velocity limits.

\input{sections/floats/fig_reachmap}

\noindent\textbf{Hand retargeting and initial pose.}
We retarget the human hand pose at every frame to a floating robot hand, as in SPIDER~\cite{spider}.
Since robot and human hands differ in shape, the retargeted hand can start penetrating the support surface or an object, a penetration the simulator cannot recover from.
We remove it with the smallest first-frame change of the wrist pose and finger joints, and a short lead-in blends this change into the original motion (\cref{sec:supp_hand}).

\noindent\textbf{Scene placement.}
The placement $\boldsymbol{\xi}=(x,y,z,\psi)$ is a translation $(x,y,z)$ and a yaw $\psi$ about the vertical axis of the demonstrated hands and objects relative to the robot base, and the support surface moves with them in height.
At a placement, each active hand's retargeted wrist pose in the robot base frame gives its arm a \emph{wrist target} at every frame.
We discretize $\boldsymbol{\xi}$ on a grid, and each placement on it is a \emph{candidate} of the clip (\cref{sec:supp_candidates}).
Many candidates are unreachable or make an arm penetrate the support surface (\cref{fig:pipeline}).
We therefore find a candidate that the arms can follow in two steps, a cheap screening that discards most candidates and an arm IK that tests the survivors.

\noindent\textbf{Screening.}
The screening reads a \emph{reachability map}~\cite{capabilitymap}, precomputed per robot hand and arm, which stores several IK solutions for wrist poses on a grid of positions and orientations (\cref{fig:reachmap}a, b).
A frame is \emph{covered} for an arm if a solution stored for a nearby wrist pose keeps the arm's and hand's lowest points near or above the support surface (\cref{fig:reachmap}c, green).
A candidate survives if at least 80\% of its frames are covered for each active arm and its objects stay within the support area (\cref{sec:supp_screening}).
Each check is a fast map lookup, and only the survivors go through the arm IK, which solves every frame (\cref{sec:exp_data}).

\noindent\textbf{Arm IK.}
An arm can reach a wrist target on several IK branches.
IK solved at each frame alone can switch branches, making the arm jump (\cref{fig:coarse}a, gray).
IK solved from the previous frame stays continuous but keeps the first frame's branch, which can later hit a joint limit or the support surface.
We therefore first choose a branch over each survivor's whole clip.
At sparse covered frames, dynamic programming picks the sequence of covering solutions with the least joint motion~\cite{stampede}, avoiding branch switches.
Linear interpolation between the picks gives the \emph{coarse path}, a configuration for every frame that only approximates the wrist targets (\cref{fig:coarse}a, navy dots, and \cref{sec:supp_coarse}).

\input{sections/floats/fig_coarse}

The arm IK then solves each frame from the previous solution, with one run starting from the coarse path's first configuration (\cref{fig:coarse}a, green).
As the sparse picks can still choose a failing branch, 31 runs start from random configurations (red).
Including the coarse path among the 32 starts, rather than 32 random ones, raises the share of kept clips from 70\% to 89\% among 64 clips (\cref{sec:supp_abl_arm}).
A run that drifts over 1\,cm from its target, \eg, at a joint limit, re-solves that frame from the coarse path and from perturbations of its previous solution.
It keeps the new solution only if it is closer without a jump.
A run passes if it stays within 3\,cm of its wrist targets on 99\% of frames, without jumps or penetration (\cref{sec:supp_armik}).
Each active arm keeps its passing run whose elbow swivel changes least, avoiding fast elbow sweeps (\cref{fig:coarse}a).
The survivor passes if these runs also avoid each other.
We adopt a passing survivor as the clip's placement (\cref{sec:supp_order}) and discard clips with none (\cref{fig:funnel}b).
Finally, after small arm and hand adjustments (\cref{sec:supp_adjust}), retiming stretches each interval until its fastest joint stays slightly below its velocity limit (\cref{fig:coarse}b, \cref{sec:supp_retime}).
The retimed clip is the kinematic reference.

%% file: sections/floats/fig_reachmap.tex
\begin{figure}[t]
  \centering
  \includegraphics[width=\linewidth]{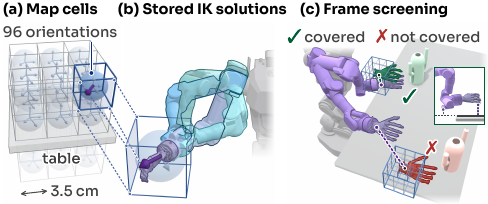}
  \caption{Reachability map. (a)~Map cells, 96 orientations each. (b)~IK solutions stored for the purple pose. (c)~Screening: a frame is covered (green) if a stored pose (purple) lies in the $3\times3\times3$ cells around its target, and uncovered (red) otherwise.}
  \aftercap{-1.8mm}
  \label{fig:reachmap}
\end{figure}

%% file: sections/floats/fig_coarse.tex
\begin{figure}[t]
  \centering
  \includegraphics[width=\linewidth]{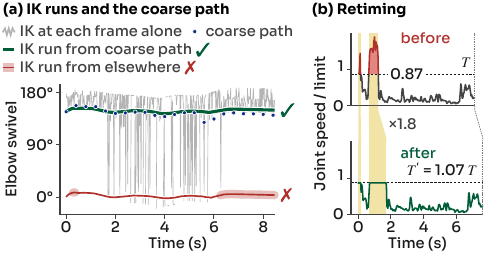}
  \caption{IK and retiming. (a)~Elbow swivel: per-frame IK (gray), coarse path (navy), runs from it (green, passes) and from random configurations (red; highlighted: wrist $>$3\,cm off). (b)~Joint speed over limit; spans above $1/1.15$ (dashed; \cref{sec:supp_retime}) stretch (yellow).}
  \aftercap{-3.0mm}
  \label{fig:coarse}
\end{figure}

%% file: sections/33_method_rl.tex
\subsection{Physics correction with residual RL}
\label{sec:rl}
We train a residual policy~\cite{maniptrans,oath,dexman} that corrects the kinematic reference in simulation~\cite{isaaclab} (\cref{fig:pipeline}), one policy per bundle of up to 64 clips.
It corrects the wrist targets and finger joints, and IK moves the arms accordingly.

\noindent\textbf{Residual action and observation.}
At each step, the reference advances one frame, and the policy gives each hand it controls a residual $\mathbf{u}_{t}=(\mathbf{u}^{p}_{t},\mathbf{u}^{R}_{t},\mathbf{u}^{f}_{t})$ for wrist position, wrist rotation, and finger joints.
The wrist residual corrects the wrist target,
\begin{equation}
\scalebox{0.92}{$\displaystyle
\mathbf{p}^{*}_{t}=\hat{\mathbf{p}}_{t+\ell}+\delta_{p}\,\mathrm{sat}(\mathbf{u}^{p}_{t}),\ 
\mathbf{R}^{*}_{t}=\hat{\mathbf{R}}_{t+\ell}\,\exp\big(\delta_{R}\,\mathrm{sat}(\mathbf{u}^{R}_{t})\big)$},
\label{eq:wrist}
\end{equation}
where $\hat{\mathbf{p}}_{t},\hat{\mathbf{R}}_{t}$ are the position and rotation of the wrist target at frame $t$, $\delta_{p},\delta_{R}$ scale the residual per axis, $\mathrm{sat}$ clips it to $[-1,1]$, and the look-ahead $\ell$ compensates for the tracking lag of the arm controller.
Differential IK turns the corrected wrist pose into arm joint targets, and the finger residual corrects the current reference finger angles.
For each hand, the policy observes its manipulated object's state, its own state and their reference errors, mostly wrist-relative (\cref{sec:supp_obs}).

\input{sections/floats/fig_funnel}\noindent\textbf{Training.}
PPO~\cite{ppo} trains each bundle from scratch, with a separate actor and value head per hand so that each hand is credited with its own reward.
Each hand's reward combines object and hand tracking, contact, force, and regularization terms (\cref{tab:supp_rewards}), of which the contact terms matter most (\cref{tab:supp_ablation}).
In both training and evaluation, a rollout ends early once a manipulated object strays from every reference pose in a short window around the current frame (\cref{sec:supp_termination}).

\noindent\textbf{Arm attachment.}
We keep the action in wrist space~\cite{dexman} rather than in arm joint space~\cite{oath}, since the wrist has fewer degrees of freedom and lies closer to the task space of the targets (\cref{tab:abl_rl}).
The policy's observations and rewards use the wrist targets, and no arm joint target enters them.
A global transformation of the scene thus applies directly to the wrist targets, and the policy runs at new placements without retraining, which enables pose augmentation (\cref{sec:data}).
A wrist pose, however, does not fix the arm's configuration.
IK solved at each reset can place the arm on a different branch, making the arm motion noisy across resets and hindering training.
We therefore reset the arm to the precomputed kinematic reference~\cite{oath,spider}, which stays on one IK branch.
With resets at random frames~\cite{deepmimic}, not only at the first, the policy trains on every phase of a clip with one consistent arm motion.
Further arm details, such as a low-pass filter before IK, are in \cref{sec:supp_action}.

\input{sections/floats/tab_e123}
\noindent\textbf{Clip acceptance.}
At the end of training, we run a batch of rollouts per clip in parallel from its first frame.
A clip is accepted for data recording (\cref{sec:data}) if at least 10\% of them reach the clip end without termination (\cref{fig:funnel}b).

%% file: sections/floats/fig_funnel.tex
\begin{figure}[t]
\includegraphics[width=\linewidth]{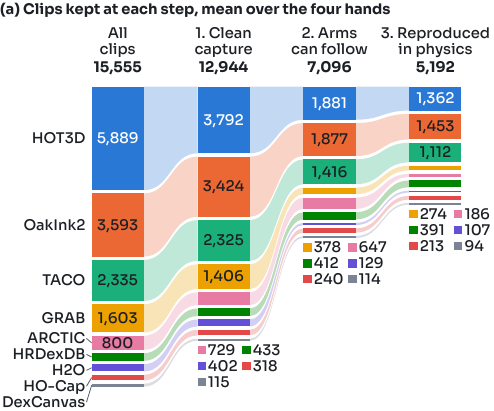}\par\vspace{5pt}
\includegraphics[width=\linewidth]{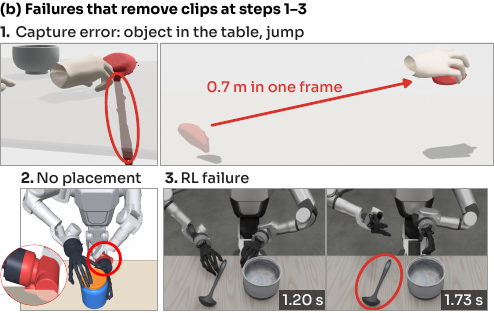}
\caption{From human sequences to robot data. (a) Clips passing data preparation (1), reference generation (2) and physics correction (3), by dataset, averaged over the four hands. (b) Failures that remove clips at each step (red).}
\aftercap{-2.3mm}
\label{fig:funnel}
\end{figure}

%% file: sections/floats/tab_e123.tex
\begin{table*}[t]
\centering
\begin{minipage}[c]{0.562\linewidth}\centering
\includegraphics[width=\linewidth]{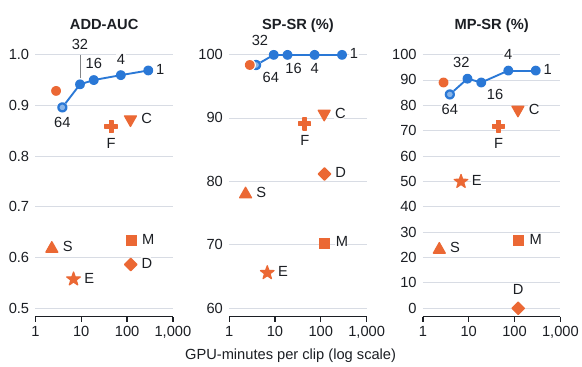}
\end{minipage}\hfill
\begin{minipage}[c]{0.426\linewidth}\centering
\fitwidth{\footnotesize\setlength{\tabcolsep}{2pt}\input{figs_tabs/tables/tab_e123_body}}
\end{minipage}
\caption{Pipelines on the same 64 TACO clips with the SharpA hand, where $K$ clips share one policy. Left: ADD-AUC, SP-SR and MP-SR against GPU-minutes per clip (log scale), each method marked by the symbol before its name. Orange marks the floating hand and blue the arm-mounted hand. Circles are ours, and light fill marks our dataset's setting. Bold: best. Method changes: \cref{sec:supp_cmp_conditions}.}
\aftercap{-1.9mm}
\label{tab:same64}
\end{table*}

%% file: figs_tabs/tables/tab_e123_body.tex
\begin{tabular}{@{}lrrrrr@{}}
\toprule
Method & $K$ & GPU-min & ADD-AUC & SP-SR & MP-SR \\
\midrule
\multicolumn{6}{l}{\makebox[8pt][c]{\textcolor[HTML]{EB6834}{\rule[0.35ex]{7pt}{1.4pt}}}\,\textit{Floating hand}} \\
\makebox[8pt][c]{\textcolor[HTML]{EB6834}{$\blacksquare$}}\,ManipTrans~\cite{maniptrans} & 1 & 122.7 & 0.63 & 70.3 & 26.6 \\
\makebox[8pt][c]{\textcolor[HTML]{EB6834}{$\blacklozenge$}}\,DexMachina~\cite{dexmachina} & 1 & 120.0 & 0.59 & 81.2 & 0.0 \\
\makebox[8pt][c]{\textcolor[HTML]{EB6834}{$\blacktriangle$}}\,SPIDER~\cite{spider} & 1 & \textbf{2.3} & 0.62 & 78.1 & 23.4 \\
\makebox[8pt][c]{\textcolor[HTML]{EB6834}{$\blacktriangledown$}}\,CHORD~\cite{chord} & 1 & 118.4 & 0.87 & 90.6 & 78.1 \\
\makebox[8pt][c]{\textcolor[HTML]{EB6834}{$\bigstar$}}\,EgoEngine~\cite{egoengine} & 1 & 6.8 & 0.56 & 65.6 & 50.0 \\
\makebox[8pt][c]{{\definecolor{mk}{HTML}{EB6834}\definecolor{mkl}{HTML}{F4AC8F}\tikz[baseline=-0.6ex]{\draw[line width=1.5pt,draw=mk] (-2.7pt,0)--(2.7pt,0) (0,-2.7pt)--(0,2.7pt);}}}\,FlashDexRetarget~\cite{flashdexretarget} & 64 & 44.7 & 0.86 & 89.1 & 71.9 \\
\makebox[8pt][c]{{\definecolor{mk}{HTML}{EB6834}\definecolor{mkl}{HTML}{F4AC8F}\tikz[baseline=-0.6ex]{\fill[mk] (0,0) circle (2pt);}}}\,DexEngine (Ours) & 64 & 2.8 & 0.93 & 98.4 & 89.1 \\
\midrule
\multicolumn{6}{l}{\makebox[8pt][c]{\textcolor[HTML]{2A78D6}{\rule[0.35ex]{7pt}{1.4pt}}}\,\textit{Arm-mounted hand (RB-Y1)}} \\
\makebox[8pt][c]{{\definecolor{mk}{HTML}{2A78D6}\definecolor{mkl}{HTML}{8AB5E8}\tikz[baseline=-0.6ex]{\draw[line width=0.6pt,draw=mk,fill=mkl] (0,0) circle (1.9pt);}}}\,DexEngine (Ours, data gen.) & 64 & 3.9 & 0.90 & 98.4 & 84.4 \\
\makebox[8pt][c]{{\definecolor{mk}{HTML}{2A78D6}\definecolor{mkl}{HTML}{8AB5E8}\tikz[baseline=-0.6ex]{\fill[mk] (0,0) circle (2pt);}}}\,DexEngine (Ours) & 32 & 9.4 & 0.94 & \textbf{100.0} & 90.6 \\
\makebox[8pt][c]{{\definecolor{mk}{HTML}{2A78D6}\definecolor{mkl}{HTML}{8AB5E8}\tikz[baseline=-0.6ex]{\fill[mk] (0,0) circle (2pt);}}}\,DexEngine (Ours) & 16 & 18.8 & 0.95 & \textbf{100.0} & 89.1 \\
\makebox[8pt][c]{{\definecolor{mk}{HTML}{2A78D6}\definecolor{mkl}{HTML}{8AB5E8}\tikz[baseline=-0.6ex]{\fill[mk] (0,0) circle (2pt);}}}\,DexEngine (Ours) & 4 & 73.0 & 0.96 & \textbf{100.0} & \textbf{93.8} \\
\makebox[8pt][c]{{\definecolor{mk}{HTML}{2A78D6}\definecolor{mkl}{HTML}{8AB5E8}\tikz[baseline=-0.6ex]{\fill[mk] (0,0) circle (2pt);}}}\,DexEngine (Ours) & 1 & 290.0 & \textbf{0.97} & \textbf{100.0} & \textbf{93.8} \\
\bottomrule
\end{tabular}

%% file: sections/34_method_data.tex
\subsection{Train-ready data generation}
\label{sec:data}

From each accepted clip, we record training data: joint states, actions, object poses, contact forces, head and exocentric images, and a language instruction.
We augment it by rerunning each policy at new placements (pose augmentation) and rendering it under five appearances (visual augmentation), yielding up to 20 episodes per clip.

\noindent\textbf{Data recording.}
We rerun each accepted clip and keep the completed rollout whose manipulated objects end closest to their reference poses (\cref{sec:supp_recording}).
This rollout is a \emph{trajectory}.
Its joint positions form $\mathcal{Q}$ of \cref{eq:goal}, its joint targets are the actions, and 
it records the normal contact force on each hand link.
We render it from egocentric stereo cameras with depth, matching the real robot's head camera, and from an exocentric camera (\cref{fig:pipeline,sec:supp_rendering}).
A vision--language model writes each clip's instruction from frames of its motion phases and a motion summary (\cref{sec:supp_language}).

\noindent\textbf{Data augmentation.}
A placement moves the objects and the hand reference together, keeping the hand--object interaction while changing the arm motion.
At a new placement, we rerun the arm IK and retiming, and the policy runs without retraining (\cref{sec:rl}).
We draw new placements from the screening survivors of \cref{sec:reference} and accept one if a completed rollout passes our penetration tests.
We keep a few accepted placements far from the adopted one and each other (\cref{sec:supp_augmentation}).
The best rollout at each kept placement becomes a trajectory, rendered under five appearances, four of them with random materials, lighting, table, and distractors.
An episode is one trajectory under one appearance.

%% file: sections/40_exp_dataset.tex
\section{Experiments}
\label{sec:experiments}
We test three claims.
(1)~The pipeline turns human demonstrations into physically executable robot data at scale, more accurately and at lower cost than prior RL-based pipelines.
(2)~Fine-tuning visuomotor policies needs physically executable hand--object interaction.
(3)~Mid-training on all of our data for one embodiment improves policies at the same fine-tuning budget, also on unseen placements.
\Cref{sec:exp_data,sec:exp_policy,sec:exp_mid} test them in turn.
Experiments with arms use an RB-Y1, a bimanual robot with 7-DoF arms on a fixed torso.

\subsection{Executable robot data at scale}
\label{sec:exp_data}
\input{sections/floats/fig_qual_sim}
\noindent\textbf{Dataset.}
\Cref{fig:funnel} shows the clips that each stage keeps (per hand: \cref{tab:supp_funnel}), \cref{fig:pipeline} what the data hold, and row~3 of \cref{fig:qual_sim} a sample recorded clip from each source dataset.
Kinematic reference generation mostly removes clips with no placement in which the fixed-base arms can follow the motion.
In GRAB, for example, a wrist travels at least 2\,m in 97\% of clips.
Physics correction removes the clips in which the hands fail to grasp and move the objects as demonstrated.
Recorded at up to four placements and in five appearances,
the 20,768 accepted clips of the four hands (37 h) yield about 65,000 trajectories (116 h) and 325,000 episodes (578 h).
With these 116 hours of trajectories, ours is the largest physics-based dataset with arms among human-to-robot pipelines, and the only one with two arms (\cref{tab:datacmp}).
We release the dataset\releasewhen, or, where a source dataset's license forbids redistribution, \mbox{the code that regenerates it}.

%% file: sections/floats/fig_qual_sim.tex
\begin{figure*}[t]
  \centering
  \includegraphics[width=\linewidth]{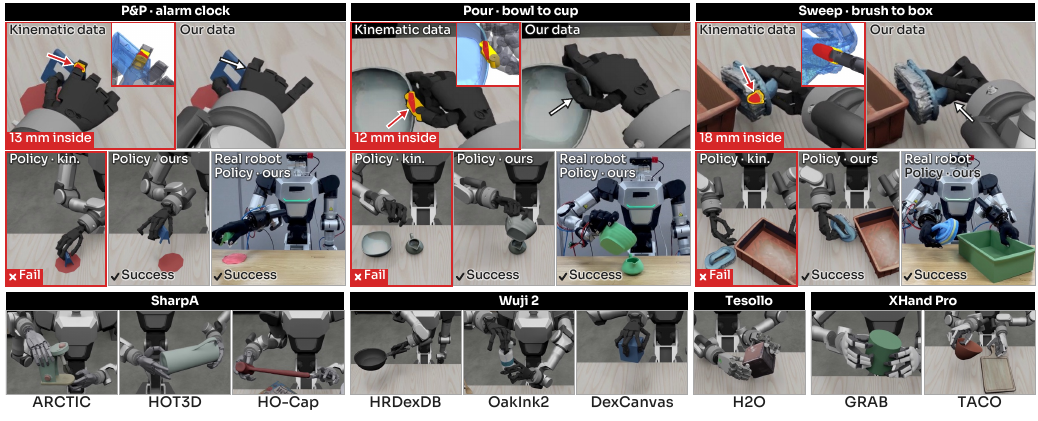}
  \vspace{-8mm}
  \caption{Data and policies. Row 1: one Wuji~2 training data frame per task in kinematic data and in ours, close to the hand, from the scene named above; arrows mark the finger link the kinematic reference pushes deepest into the object (yellow; red: its part inside the object; inset: the same frame, object see-through). Row 2: GR00T N1.7 trained on kinematic replay and on ours, in simulation at the same time step of that scene, and the policy trained on ours on the real robot. Row 3: our data, one clip per source dataset, hand named above.}
  \label{fig:qual_sim}
\end{figure*}

%% file: sections/41_exp_pipeline.tex
\noindent\textbf{Comparison with existing methods.}
With a fixed seed, we randomly draw 64 of CHORD's released two-handed TACO clips~\cite{chord} that pass data preparation and kinematic reference generation.
We score all methods with CHORD's object-tracking evaluation.
ADD-AUC integrates object-vertex accuracy over thresholds from 1 to 9\,cm.
SP-SR and MP-SR are success rates under the object criteria of SPIDER~\cite{spider} and ManipTrans~\cite{maniptrans}.
Methods that train a policy are scored by the best of 64 rollouts per clip, as our data recording keeps one of at least 64 rollouts.
SPIDER and EgoEngine output one trajectory per run, scored without selection.
Cost is GPU-minutes per clip on an NVIDIA L40S and counts only RL training for methods that train a policy.
We run ours both on a floating hand with six virtual wrist joints, as prior methods do, and with the arm attached, where we vary the number of clips $K$ per policy.

On the floating hand (\cref{tab:same64}), our policy shared by all 64 clips (\mkfloat) reaches ADD-AUC 0.93 at 2.8 GPU-minutes per clip.
CHORD reaches 0.87 at 118, and FlashDexRetarget, which also shares one policy, 0.86 at 44.7.
With the arm attached, one policy per clip is most accurate, ADD-AUC 0.97 at 290 GPU-minutes (\mkarmk, $K=1$).
At $K=64$, our dataset's setting (\mkarm), the cost falls about 75-fold to 3.9 GPU-minutes while ADD-AUC drops only to 0.90, still above every prior method.
Training with floating hands and attaching the arm later, even at our placements, passes 26\% fewer clips through our arm IK,
as arm-free RL sometimes moves the wrist where the arm cannot follow (\cref{sec:supp_posthoc}).

\input{sections/floats/tab_abl_rl}
\noindent\textbf{Ablations.}
Our placement search keeps a kinematic reference for 43.8\% of 128 sampled clips, against 1.6\% at the default scene position (\cref{sec:supp_abl_place}).
Screening thus buys speed, not clips: testing every candidate keeps 45.3\%, a superset of ours, but takes three times longer.
A single IK run that solves each frame from the previous one, starting at the rest pose and without the coarse path, re-solving or the choice among runs, keeps only 28\% of the 64 clips within the arm limits, against 89\% for our arm IK (\cref{sec:supp_abl_arm}).

In the RL ablation (\cref{tab:abl_rl}), each run trains the 64 clips of \cref{tab:same64} as one bundle.
Without a precomputed arm motion, a random start has only its frame's wrist pose to solve the arm from.
Such solutions move some arm joint by more than 20$^\circ$ between adjacent frames in 43\% (right) and 32\% (left) of frame pairs, which never occurs on our reference.
With residuals on the arm joints, resetting the arm to our reference raises ADD-AUC from 0.544 to 0.856.
With residuals on the wrist, it raises ADD-AUC from 0.832 to 0.900, the highest.
With wrist residuals, re-solving the arm scores even below training without random starts (0.888).
These results suggest that residuals are easier to learn on the wrist, which lies closer to the task space of the targets.
They also suggest that random starts help only when they place the arm on one consistent motion.
\Cref{sec:supp_abl_rl} ablates the other elements of the recipe, such as the contact terms.

%% file: sections/floats/tab_abl_rl.tex
\begin{table}[t]
\centering
{\footnotesize\setlength{\tabcolsep}{2pt}
\input{figs_tabs/tables/tab_abl_rl_body}}
\caption{Arm attachment in RL on the 64 clips of \cref{tab:same64} (SharpA hand, $K=64$, 1,600 iterations). Residuals act on the arm joints (1) or the wrist pose (2), and each random start re-solves the arm by IK or resets it to our kinematic reference. Best of 64 rollouts, early termination on (unlike \cref{tab:same64}). More changes in \cref{tab:supp_ablation}.}
\aftercap{-3mm}
\label{tab:abl_rl}
\end{table}

%% file: figs_tabs/tables/tab_abl_rl_body.tex
\begin{tabular*}{\columnwidth}{@{\extracolsep{\fill}}lrrr@{}}
\toprule
Training & ADD-AUC & SP-SR & MP-SR \\
\midrule
\multicolumn{4}{@{}l}{\textit{(1) Arm-joint residuals}} \\
\hspace*{4pt}Arm re-solved by IK at random starts & 0.544 & 25.0 & 20.3 \\
\hspace*{4pt}Arm reset to our reference & 0.856 & 76.6 & 73.4 \\
\midrule
\multicolumn{4}{@{}l}{\textit{(2) Wrist-space residuals}} \\
\hspace*{4pt}Arm re-solved by IK at random starts & 0.832 & 75.0 & 71.9 \\
\hspace*{4pt}No random starts (first frame only) & 0.888 & 82.8 & 79.7 \\
\hspace*{4pt}Arm reset to our reference (Ours) & 0.900 & 85.9 & 81.2 \\
\bottomrule
\end{tabular*}

%% file: sections/42_exp_policy.tex
\subsection{Physics correction improves policies}
\label{sec:exp_policy}
\input{sections/floats/tab_e5}
We test whether contact physics matters for fine-tuning visuomotor policies, even when data without it have plausible images and action labels.
To do so, we remove only the physics correction: \emph{kinematic replay} data render our kinematic reference for the same clips and placements, with the reference joint poses as action labels and the objects on their ground-truth trajectories.
In them, fingers can penetrate objects and contact dynamics are absent, unlike in our physically corrected data (row~1 of \cref{fig:qual_sim}).

\noindent\textbf{Setup.}
We use three tasks, pick-and-place (P\&P), Pour and Sweep, each with 20 Wuji~2 clips from our generated data.
For these tasks, we run the augmentation of \cref{sec:data} more densely, with up to 100 placements per clip, each rendered with up to 20 visual augmentations.
All sources share these clips and placements.
We train two policies: Diffusion Policy (DP)~\cite{diffusionpolicy} from scratch and GR00T N1.7~\cite{gr00t} fine-tuned from public weights (\cref{sec:supp_policy_training}).

\noindent\textbf{Results.}
We test two levels of placement: L0 uses five training placements per clip, and L1 five held-out placements of the same clip.
Both hold only scenes where replaying our recorded actions succeeds, and every test scene is thus physically executable.
Success requires the object lifted and at its target in P\&P, the demonstrated pouring pose in Pour, and tool travel in contact with the target in Sweep (\cref{sec:supp_success}).
\Cref{tab:main} shows that without physics correction, policies mostly fail: DP succeeds in 0\% of rollouts at L0 and 1\% at L1, against 53\% and 40\% on our data.
GR00T N1.7 succeeds in 6\% and 5\%, against 65\% and 50\%, even with 3.5 to 5 times as many fine-tuning steps.
Plausible images and action labels are thus not enough, and fine-tuning needs hands that actually grasp the objects.
Row~2 of \cref{fig:qual_sim} shows one scene per task where the policy succeeds only with our data.
Both policies succeed in at most 5\% of rollouts on data from SPIDER~\cite{spider}, recorded in its own MuJoCo simulation, given the arm by its own rule (its Sec.~A.6), and kept in full.
It tracks objects less accurately than ours (\cref{tab:same64}), and even on successful clips, the attached arm misses its wrist by over 3\,cm in 28\% of trajectories.

%% file: sections/floats/tab_e5.tex
\begin{table}[t]
\centering
{\footnotesize\setlength{\tabcolsep}{0pt}
\input{figs_tabs/tables/tab_e5_body}}
\caption{Simulated success (\%), Wuji~2 hands, 100 scenes per task and level. L0: training placements, L1: held out. + mid: mid-trained first. GR00T steps: 40,000 on SPIDER + arm and ours, 140,000--199,000 ($\approx$150 passes) on kinematic replay. Bold: best.}
\aftercap{-1.9mm}
\label{tab:main}
\end{table}

%% file: figs_tabs/tables/tab_e5_body.tex
\begin{tabular*}{\linewidth}{@{\extracolsep{\fill}}l*{3}{r@{\extracolsep{0pt}\hspace{7pt}}r@{\extracolsep{\fill}}}@{\hspace{5pt}}r@{\extracolsep{0pt}\hspace{7pt}}r@{}}
\toprule
 & \multicolumn{2}{c}{P\&P} & \multicolumn{2}{c}{Pour} & \multicolumn{2}{c}{Sweep} & \multicolumn{2}{c}{Mean} \\
\cmidrule(lr){2-3} \cmidrule(lr){4-5} \cmidrule(lr){6-7} \cmidrule(l){8-9}
Policy & L0 & L1 & L0 & L1 & L0 & L1 & L0 & L1 \\
\midrule
\multicolumn{9}{@{}l}{\textit{Kinematic replay data}} \\
\hspace*{6pt}DP~\cite{diffusionpolicy} & 0 & 0 & 0 & 2 & 0 & 1 & 0 & 1 \\
\hspace*{6pt}GR00T~\cite{gr00t} & 12 & 8 & 3 & 4 & 3 & 3 & 6 & 5 \\
\midrule
\multicolumn{9}{@{}l}{\textit{SPIDER}~\cite{spider} \textit{+ arm data}} \\
\hspace*{6pt}DP & 0 & 1 & 0 & 0 & 4 & 3 & 1 & 1 \\
\hspace*{6pt}GR00T & 4 & 5 & 2 & 1 & 2 & 2 & 3 & 3 \\
\midrule
\multicolumn{9}{@{}l}{\textit{DexEngine data (Ours)}} \\
\hspace*{6pt}DP & 32 & 32 & 58 & 39 & 70 & 48 & 53 & 40 \\
\hspace*{6pt}GR00T & 50 & 45 & 74 & 52 & 70 & 54 & 65 & 50 \\
\hspace*{6pt}GR00T + mid & \textbf{72} & \textbf{48} & \textbf{75} & \textbf{61} & \textbf{81} & \textbf{62} & \textbf{76} & \textbf{57} \\
\bottomrule
\end{tabular*}

%% file: sections/43_exp_real.tex
\subsection{Mid-training and the real robot}
\label{sec:exp_mid}
\input{sections/floats/tab_real}
\noindent\textbf{Mid-training.}
To use our data's scale and diversity, we mid-train GR00T on all our embodiment-specific data, excluding the source sequences of the task clips, and then fine-tune it as in \cref{sec:exp_policy} (\cref{sec:supp_policy_training}).
With the same 40K fine-tuning steps, mid-training raises success from 65\% to 76\% at L0 and from 50\% to 57\% at L1 (\cref{tab:main}).
Longer fine-tuning instead fits the training placements, raising L0 to 82\% but leaving L1 at 50\% (55\% with mid-training, \cref{tab:supp_e5_long}), which suggests that mid-training mainly helps policies generalize to held-out placements.

\noindent\textbf{Real robot.}
We then test whether policies fine-tuned only on our simulated data work on a real robot.
We deploy GR00T and GR00T + mid on an RB-Y1 with two Wuji~2 hands (row~2 of \cref{fig:qual_sim}, right).
We test one clip per task, in five trials at fine-tuning placements (L0) and five at held-out ones (L1).
Both policies are fine-tuned on these clips from the public and the mid-trained weights, with strong visual augmentation (\cref{sec:supp_real}).
To assess how well the simulated trajectories transfer to the real robot, we also replay each clip’s recorded trajectory open loop.
The replay succeeds in 87\% of trials at L0 and 80\% at L1 (\cref{tab:real}).
The recorded trajectories thus largely transfer to the real robot.
GR00T + mid succeeds in 73\% at L0 and 27\% at L1, against 20\% and 0\% without it.
Mid-training also helps on the real robot, where the policy without it mostly fails.

%% file: sections/floats/tab_real.tex
\begin{table}[t]
\centering
{\footnotesize\setlength{\tabcolsep}{4pt}
\input{figs_tabs/tables/tab_real_body}}
\caption{Real-robot success (\%) in five trials per task and level, one clip per task (\cref{sec:supp_real}), Wuji~2 hands, no real training data. L0: fine-tuning placements, L1: held out. Replay: recorded trajectory played open loop. Bold: best policy.}
\aftercap{-6mm}
\label{tab:real}
\end{table}

%% file: figs_tabs/tables/tab_real_body.tex
\begin{tabular*}{\linewidth}{@{\extracolsep{\fill}}l*{8}{r}@{}}
\toprule
 & \multicolumn{2}{c}{P\&P} & \multicolumn{2}{c}{Pour} & \multicolumn{2}{c}{Sweep} & \multicolumn{2}{c}{Mean} \\
\cmidrule(lr){2-3}\cmidrule(lr){4-5}\cmidrule(lr){6-7}\cmidrule(lr){8-9}
 & L0 & L1 & L0 & L1 & L0 & L1 & L0 & L1 \\
\midrule
Replay & 80 & 80 & 80 & 60 & 100 & 100 & 87 & 80 \\
\midrule
GR00T~\cite{gr00t} & 20 & 0 & 40 & 0 & 0 & 0 & 20 & 0 \\
GR00T + mid & \textbf{60} & 0 & \textbf{80} & \textbf{20} & \textbf{80} & \textbf{60} & \textbf{73} & \textbf{27} \\
\bottomrule
\end{tabular*}

%% file: sections/50_conclusion.tex
\section{Conclusion}
\label{sec:conclusion}
{\looseness=-1
We presented DexEngine, a data pipeline that turns human hand--object demonstrations into physically executable, train-ready data for a whole bimanual robot.
Searching each clip's placement and arm motion before physics keeps the clips the arms can follow.
Acting in wrist space and resetting the arm to this motion at random starts let one policy correct up to 64 clips with the arms attached, more accurately than prior floating-hand methods.
Policies fine-tuned on kinematic replays of the same clips mostly fail, showing that fine-tuning needs physically accurate hand--object interaction.
Mid-training on our data helps further, also on a real robot.\par}

\noindent\textbf{Limitations.}
Physics correction with closed-loop RL remains costly.
Human captures also transfer less often than expected.
About a third of our clips become robot data (\cref{fig:funnel}), and Do as I Do~\cite{doasido} keeps only 4\% of sampled internet videos after reconstruction.
Apart from capture defects, most lost clips have no placement the arms can follow from a fixed torso and base (\cref{sec:exp_data}).
Moving the torso and base, or capture guidelines that respect the robot's limits, could make more human data usable.

%% file: common/supp.tex
\appendix
\setcounter{table}{0}\setcounter{figure}{0}\setcounter{equation}{0}
\renewcommand{\thetable}{S\arabic{table}}\renewcommand{\thefigure}{S\arabic{figure}}\renewcommand{\theequation}{S\arabic{equation}}
\providecommand{\theHtable}{}\providecommand{\theHfigure}{}\providecommand{\theHequation}{}
\renewcommand{\theHtable}{S\arabic{table}}\renewcommand{\theHfigure}{S\arabic{figure}}\renewcommand{\theHequation}{S\arabic{equation}}
This supplement follows the main paper.
\Cref{sec:supp_data,sec:supp_reference,sec:supp_details,sec:supp_generation} detail the four stages of \cref{sec:parsing,sec:reference,sec:rl,sec:data}.
\Cref{sec:supp_experiments} gives the setups and further results of \cref{sec:experiments}, and \cref{sec:supp_gallery} shows samples of the recorded data.

\input{sections/S1_supp_data}

\input{sections/S2_supp_reference}
\input{sections/S3_supp_rl}
\input{sections/S4_supp_generation}
\input{sections/S5_supp_experiments}
\input{sections/S6_supp_gallery}

%% file: sections/S1_supp_data.tex
\section{Data preparation details}
\label{sec:supp_data}
We use the hand poses (MANO~\cite{mano} parameters), object poses and object meshes of nine public datasets~\cite{hot3d,oakink2,taco,grab,arctic,hrdexdb,h2o,hocap,dexcanvas}.
All sequences are resampled to 50\,Hz.

\subsection{Scene frame, table and supports}
\label{sec:supp_layout}
\noindent\textbf{Frame alignment.}
Each dataset is converted to a z-up frame and rotated about the vertical axis so that the mean direction from the MANO wrist to the fingertips matches that of TACO.
HOT3D is aligned per clip instead, since its hand direction varies widely between clips.

\noindent\textbf{Table height.}
The \emph{support surface} of the main text is the static table on which the objects rest.
ARCTIC has none, since its hands often pass below the surface the objects rest on.
The sequence defines the table plane in the tabletop datasets (TACO, HO-Cap, DexCanvas, HRDexDB) and in the raw motion-capture frame of OakInk2.
For GRAB and H2O, we set the table at the height where objects rest, \ie, stay still for at least 0.5\,s with no fingertip within 2.5\,cm.
For HOT3D, we assume the table is the lowest surface on which a handled object rests, within the clip or within 2\,s of it.
Handled objects are the manipulated objects and every object that moves in the clip.

\noindent\textbf{Supports.}
An unheld object needs a support when it rests at least 1\,cm above the table, and whenever it rests in ARCTIC.
A tracked object directly below it becomes a \emph{tracked support}, which collides with the robot and the objects.
Otherwise, a disc covering the object's footprint stands under it for the whole clip.
Since the real support is not observed and a disc can be larger than it, discs collide only with objects and cannot block the robot hand.
In the rendered data, each disc stands on a pillar down to the table or the floor.

\subsection{Hand--object assignment}
\label{sec:supp_assignment}
All rules read the MANO hand and the object poses.
A hand \emph{grasps} an object when at least two fingertips (one, for objects whose largest extent is below 8\,cm) are within 2.75\,cm of its surface or the palm touches it.
It \emph{carries} the object when the object surface moves at 3\,cm/s or more together with the fingertips, when the object rotates in place, or when the hand drives its articulation.
A hand that grasps for at least 0.5\,s in total manipulates the object it carries longest, if that carry lasts at least 0.3\,s, and otherwise the one it grasps longest (at least 0.3\,s).
A hand is \emph{active} if it manipulates an object.
The arm of an inactive hand holds a rest pose.
When both hands manipulate the same object, it becomes one shared object.

\subsection{Clip division}
\label{sec:supp_division}
We take the sequences as the datasets provide them, counting OakInk2 by its annotated primitive tasks (2,828 of the 7,899 sequences of all datasets, \cref{tab:supp_funnel}).
HOT3D sequences are first cut into windows in which a hand touches the same object.
In every dataset, sequences (for HOT3D, these windows) longer than 15\,s are split in the middle of an interval in which the objects are still and the hands are away, or otherwise at the slowest wrist motion, so that nearly every piece lasts at least 5\,s.
Each piece is then cut wherever a hand switches to a different object on its contact timeline. 
A switch to an object that no hand manipulates counts only if that object moves.
Handovers, and alternation between one and two hands holding the same object, stay in one clip.
At a switch, both clips include the idle gap between the two grasps.

\subsection{Clip filtering}
\label{sec:supp_filtering}
A clip is removed when
(i)~no hand manipulates an object;
(ii)~tracking fails: in HOT3D, a pose gap longer than 0.3\,s while the object moves; an object that moves by 2\,cm or more without contact, a fingertip keypoint jump, or an object faster than 3\,m/s while the wrist is slow;
(iii)~the scene is inconsistent, because the object goes more than 2\,cm below the table, a manipulated object that is not held at the start lies more than 1\,cm inside it, the wrist goes more than 5\,cm below the table, the estimated GRAB table lies more than 2\,cm from the table that GRAB provides, or, in ARCTIC, a manipulated object is neither touched by a hand nor resting on a support; or
(iv)~the clip is shorter than 1\,s.
A clip is also removed if its retargeted fingertips deviate from the human fingertips by more than 20\,mm on average.
Only this check depends on the robot hand.
\Cref{tab:supp_filters} lists the thresholds.

\input{sections/floats/tab_supp_filters}

%% file: sections/floats/tab_supp_filters.tex
\begin{table}[h]
\centering
{\footnotesize\setlength{\tabcolsep}{3pt}
\begin{tabular}{@{}l>{\raggedright\arraybackslash}p{0.712\linewidth}@{}}
\toprule
Check & Rejected if \\
\midrule
\multicolumn{2}{@{}l}{\textit{Data preparation (\cref{sec:parsing})}} \\
Manipulated object & no hand grasps for $\geq$0.5\,s in total and carries ($\geq$0.3\,s) or grasps ($\geq$0.3\,s) one object \\
\addlinespace[1.5pt]
Source tracking & object moves $\geq$2\,cm without contact; $\geq$3 frames of fingertip jumps $>$30\,mm while the wrist turns $<$5$^\circ$; moving pose gap $>$0.3\,s (HOT3D); object $>$3\,m/s while wrist $<$1.5\,m/s \\
\addlinespace[1.5pt]
Scene & object hull $>$2\,cm below table; manipulated object not held at start $>$1\,cm inside table; MANO wrist $>$5\,cm below table; table height $>$2\,cm from the table GRAB provides (GRAB); manipulated object neither touched nor supported (ARCTIC) \\
\addlinespace[1.5pt]
Length & $<$1\,s \\
\addlinespace[1.5pt]
Retargeting & mean fingertip error $>$20\,mm \\
\midrule
\multicolumn{2}{@{}l}{\textit{Kinematic reference generation (\cref{sec:reference})}} \\
Placement & no placement within 1,024 candidates / 1,200\,s (2,048 candidates for one active arm); an active arm covers $<$80\% of frames; a manipulated object starts within the 15\,cm strip next to the robot \\
\addlinespace[1.5pt]
Objects & a manipulated object, where it starts or comes to rest, or a fixed object that must stand in the scene, overlaps the area where the robot stands or has $>$10\% of its footprint beyond the table top (in the placement search and again after retiming) \\
\addlinespace[1.5pt]
Arm & kept coarse-path solutions on $<$50\% of sparse frames; wrist position p99 error $>$30\,mm; arm joint vector jump $>$1\,rad; elbow swivel $>$180$^\circ$ in 0.5\,s; arm flip without a continuous solution \\
\addlinespace[1.5pt]
Collision & arm--table, arm--fixed-object, arm/hand--torso or arm--arm penetration $>$2\,mm; at the first frame, an arm inside a manipulated object or two manipulated objects inside each other by $>$1\,cm \\
\addlinespace[1.5pt]
Timing & retiming ratio $>$3; ratio of joint speed to limit $>$1 at the 99th percentile of frames or $>$3 at any frame \\
\addlinespace[1.5pt]
Wrist and fingers & wrist position error $>$70\,mm on $>$5\% of frames in which a hand touches a moving object; finger angle $>$0.5$^\circ$ outside the joint range used in retargeting \\
\bottomrule
\end{tabular}}
\caption{Thresholds for removing clips in data preparation and in kinematic reference generation.}
\label{tab:supp_filters}
\end{table}

%% file: sections/S2_supp_reference.tex
\section{Kinematic reference generation details}
\label{sec:supp_reference}
\Cref{tab:supp_filters} lists the thresholds at which kinematic reference generation removes clips.

\subsection{Hand retargeting and initial pose}
\label{sec:supp_hand}
\noindent\textbf{Hand retargeting.}
We follow the MuJoCo~\cite{mujoco} keypoint retargeting of SPIDER~\cite{spider}.
The robot palm frame is tied to the MANO wrist pose and the five robot fingertips to the MANO fingertip positions by soft equality constraints.
Each frame is a constrained solve, warm-started from the previous frame, that approximately minimizes
\begin{equation}
\begin{aligned}
& \mathbf{q}^{h}_{t} \approx \arg\min_{\mathbf{q}}\;
\textstyle\sum_{i=1}^{5} k_{i}\,\lVert \mathbf{p}_{i}(\mathbf{q})-\hat{\mathbf{p}}_{i,t}\rVert^{2} \\
& \quad + k_{p}\,\lVert \mathbf{p}_{w}(\mathbf{q})-\hat{\mathbf{p}}_{w,t}\rVert^{2}
+ k_{R}\,\lVert \log(\hat{\mathbf{R}}_{w,t}^{\top}\mathbf{R}_{w}(\mathbf{q}))\rVert^{2} \\
& \quad + \lambda\,\lVert \mathbf{q}-\mathbf{q}^{h}_{t-1}\rVert^{2},
\end{aligned}
\label{eq:retarget}
\end{equation}
within the joint limits, at 50\,Hz, where $\mathbf{p}_{i}$ are the fingertip positions and $\mathbf{p}_{w},\mathbf{R}_{w}$ the palm position and orientation.
We try several initial wrist orientations at the first frame.
For the Tesollo and Wuji~2 hands, we remove the hyperextension range, opposite to flexion, of the finger joints from the distal joints up to the MCP joint, since backward-bent fingers give unnatural motion and enlarge the search space of RL.

\noindent\textbf{Hand tilt near the table.}
Before the initial-pose repair, we tilt the hand on frames in which the last arm link would enter the table while the palm stays above it.
This link sits where the human wrist was, and no placement can fix such frames. 
The hand is tilted about its contacting fingertips by at most 20$^\circ$ and lifted by at most 30\,mm, keeping the object motion unchanged, and the fingers are then re-solved.

\noindent\textbf{Initial hand pose.}
The initial-pose repair removes a first-frame penetration too deep for the simulation to push out, by the smallest change
$\Delta\mathbf{q}^{h}=(\Delta\mathbf{p},\Delta\boldsymbol{\theta},\Delta\mathbf{q}_{f})$
of the robot hand's pose $\mathbf{q}^{h}_{1}$ there, which comprises the floating hand's wrist position, wrist rotation and finger joints:
\begin{equation}
\begin{aligned}
\min_{\Delta\mathbf{q}^{h}}\; \lVert \mathbf{w}\odot\Delta\mathbf{q}^{h}\rVert^{2}
&+\lambda_{n}\,n_{\mathrm{pen}}(\mathbf{q}^{h}_{1}\oplus\Delta\mathbf{q}^{h}) \\
&+\lambda_{d}\,d_{\mathrm{pen}}(\mathbf{q}^{h}_{1}\oplus\Delta\mathbf{q}^{h}),
\end{aligned}
\label{eq:f0}
\end{equation}
where $\oplus$ adds the position and finger changes and composes the rotation, $\odot$ is the element-wise product, $\mathbf{w}$ is a weight vector, $n_{\mathrm{pen}}$ counts the points where the hand penetrates the table or an object, $d_{\mathrm{pen}}$ sums their depths, and $\lambda_{n},\lambda_{d}$ weight the two penalties.
The weights $\mathbf{w}$ make thumb and index changes the cheapest, then the other fingers, then the wrist rotation, then the wrist position, and 1\,cm of wrist motion costs as much as 0.6\,rad of thumb or index motion.
We use $\lambda_{n}=10^{3}$ per point and $\lambda_{d}=5\times10^{4}$ per meter, so that any penetration outweighs the change cost.
We solve \cref{eq:f0} with the derivative-free Nelder--Mead method in three stages, changing the fingers only, then the wrist only, then both, and stop at the first stage that finds a feasible solution.
A short lead-in, prepended to the clip, blends the repaired first pose into the original motion at bounded finger and wrist speed.

\subsection{Placement candidates}
\label{sec:supp_candidates}
Offsets form a search window on a 3.5\,cm grid, $\pm$42\,cm forward and $\pm$73.5\,cm sideways of its center.
The center is the \emph{default scene position}, where data preparation leaves the scene in front of the robot, with the origin of the aligned capture frame 70\,cm ahead of the robot base.
For HRDexDB, HOT3D, H2O and GRAB, whose capture frames can place a clip too far from the robot, the default scene position instead puts the mean wrist position of each clip at a fixed point per dataset, about 36\,cm ahead of the robot base.
The yaw takes 72 values in 2.5$^\circ$ steps in $[-90^\circ,90^\circ)$ about the mean wrist position of the clip.
Heights $z$ lie 7\,cm apart, with one to seven values per dataset because datasets keep the hands at different heights above the table.
ARCTIC, which has no table, uses twelve values, and neither the table tests nor the table-area test applies to it.
This gives at most $25\times43\times72\times7=541{,}800$ candidates, or $928{,}800$ for ARCTIC.

\subsection{Reachability map and screening}
\label{sec:supp_screening}
\noindent\textbf{Reachability map.}
The map has its own 3.5\,cm position grid, separate from the grid of placement offsets.
The orientations come from the retargeted wrist orientations of TACO.
Since a placement turns the scene, we copy them at seven further turns about the vertical axis, 45$^\circ$ apart, and thin them to 96 by farthest point sampling.
For each pose, IK runs from 16 starts for 60 damped least-squares iterations.
The solutions are taken in order of their distance from the joint limits, and one is stored if it reaches its pose within 5\,mm and 5$^\circ$ and differs from every stored one by at least 25$^\circ$ in some joint, up to four per pose.
The map depends on the robot hand through these orientations and the hand's lowest points.

\noindent\textbf{Screening.}
A map pose is near a wrist target if it lies in the target's cell or one of the 26 neighboring cells and its orientation is among the three map orientations closest to the target's (\cref{fig:reachmap}c), since with 96 orientations the closest one alone could miss reachable targets.
The screening lets the lowest points of the arm and the hand lie up to 10\,cm and 5\,cm below the table top, which avoids rejecting feasible candidates, and the arm IK then applies the 2\,mm limit to the arm links.
The 80\% coverage that a candidate needs for each active arm (\cref{sec:reference}) is lenient, since the coarse orientations can leave frames uncovered that the arm IK reaches.
Objects must also stay on the tabletop, and no manipulated object may start in the 15\,cm strip of the table next to the robot, since an object close to the robot makes the arm fold into the torso. 

\subsection{Candidate order}
\label{sec:supp_order}
A candidate passes when its arm IK passes (\cref{sec:supp_armik}).
The first round runs three series of at most 12 tests, one at the zero shift $(x,y)=(0,0)$, the default scene position, and two along straight lines toward it from points 35\,cm to its left and right.
A series visits the grid points of its line in order and stops at its first pass.
At each point it tests the heights $z$ of the dataset from the highest down, each with the surviving candidate closest to the point within three grid cells.
If none passes, the second round orders the survivors by their fraction of covered frames, counted without the neighboring cells and with the 2\,mm table limit of the arm IK on the arm links.
At the same yaw and height $z$, the second round skips the grid positions within two cells of an earlier pass, so that the passing candidates differ.
It also skips the neighbors of a candidate whose wrist position error is too large for them to pass, assuming the error changes no more than the wrist targets move.
It stops once three candidates pass or the budget runs out, at 1,024 candidates, 2,048 for a clip with one active arm, or 1,200\,s.
The first passing candidate becomes the clip's placement, and the others are fallbacks, tried in order if the placement fails a later step of reference generation.

\subsection{Coarse path}
\label{sec:supp_coarse}
The coarse path works on sparse frames, one every 0.4\,s.
At each sparse frame, the solutions that the map stores for the poses near the wrist target (\cref{sec:supp_screening}) are refined toward it by eight damped least-squares iterations.
Those that reach it within 1\,cm and 8$^\circ$ without entering the table beyond a lenient margin are kept, and dynamic programming picks one per sparse frame to minimize
\begin{equation}
\begin{aligned}
&\textstyle\sum_{t}\big(c(\rho_{t})+\zeta_{t}\,\bar{\rho}_{t}+d_{t}-\eta\,m_{t}\big),\\
&r_{j,t}=\frac{\lvert q^{a}_{j,t}-q^{a}_{j,t-1}\rvert}{v_{j}\,\Delta t_{s}},\quad
\rho_{t}=\max_{j} r_{j,t},\quad \bar{\rho}_{t}=\tfrac{1}{7}\textstyle\sum_{j} r_{j,t},
\end{aligned}
\label{eq:dp}
\end{equation}
where $q^{a}_{j,t}$ is arm joint $j$ at sparse frame $t$, $v_{j}$ its velocity limit, $\Delta t_{s}=0.4$\,s the interval between sparse frames, and $c(\rho)=\rho$ for $\rho\le 1$ and $50\,\rho$ otherwise.
The travel term, with $\zeta_{t}=0.5$, or 4.5 where the wrist target stands still, keeps many joints from each drifting a little while the target rests.
The term $d_{t}$ is how far, in meters, the arm or hand lies below the table top in the stored solution that the pick refines, and $m_{t}$, weighted by $\eta=10^{-3}$, rewards distance from the joint limits.
Between sparse frames, the coarse path interpolates each arm joint linearly between the picks, and a sparse frame without a kept solution is skipped, so that the interpolation spans it.
A candidate fails here, before the costly arm IK, if fewer than half of the sparse frames have a kept solution.

\subsection{Arm IK}
\label{sec:supp_armik}
\noindent\textbf{Limits of the arm IK.}
\emph{Tracking} means a wrist position error of at most 30\,mm on 99\% of the frames, with each frame weighted by the stretch that retiming would give it (\cref{eq:retime}).
The arm IK clamps every joint at its position limits.
A \emph{jump} is a change of the arm joint vector by more than 1\,rad (L2 norm) between consecutive frames, or an elbow swivel, the rotation of the elbow about the shoulder--wrist line, of more than 180$^\circ$ within 0.5\,s.
A joint turning more than half a turn within 2\,s marks an arm flip.
The motion is then replaced by a continuous solution at the same placement that turns the arm less, and the candidate fails if none within the limits is found.
\emph{Penetration} is limited to 2\,mm.
The \emph{fixed objects} are the supports and other still objects that simulation holds in place.
The table and fixed-object tests cover the arm links.
The torso test, which checks the robot body of the main text, also covers the palm and fingers, and the test between the two arms covers the arm links of both.
After the first frame, the arms are not tested against moved objects, which simulation collides with the arms.

\noindent\textbf{IK runs.}
At the first frame, 32 runs start from the first pose of the coarse path and 31 random configurations, each solved by damped least squares.
Runs that reach the same solution are merged, and up to eight distinct runs, the coarse-path one first, that reach the target within 1\,cm and 8$^\circ$ continue to the next frames.
A run that ends a frame more than 1\,cm from the target re-solves that frame from the coarse-path pose.
If it is still more than 1\,cm off, it also re-solves from 32 configurations sampled within $\pm$0.5\,rad per joint of its previous solution.
A re-solve is kept only if it lowers the error and moves the arm joints by at most 1\,rad (L2 norm) from the previous solution.
Each arm has a budget of 12 re-solved frames.
The arm IK stops early, and the candidate fails, once more than three frames and more than 1\% of the frames have every run more than 30\,mm off.
After the last frame, the up to four runs with the lowest error within the tracking limit are tested against the limits above, apart from the fixed-object and other-arm tests.
Of those that pass, the one whose elbow swivels least within any 0.5\,s is kept for each arm, and the kept runs of both arms must then pass the fixed-object and other-arm tests together.

\subsection{Adjustments after the arm IK}
\label{sec:supp_adjust}
\noindent\textbf{Arm refinement.}
Before hand separation, IK solves the lead-in, which the arm IK does not cover.
A few damped least-squares steps then refine the kept arm motion at every frame rather than solve it anew, so that it stays on the solution that passed.
If the refined motion fails the tracking, retiming or collision limits, dynamic programming over the stored solutions at every frame~\cite{stampede} solves it anew under these limits.
The candidate fails if that solution fails too.

\noindent\textbf{Hand separation.}
Hand separation treats penetrations deeper than 2\,mm between the two hands, between a hand and the other hand's object and between a hand and the torso.
It always treats the first frame if any of these pairs is closer than 2\,mm there, since the simulation starts from it.
Pairs that include an arm link are not treated, since the wrist shifts and finger spreads that this step chooses act on the hand links, and IK places the arm links.
In frames without contact, fingers caught in a penetration spread first, by at most 12$^\circ$.
The wrist then shifts smoothly over time by at most 30\,mm, in the direction that leaves the largest clearance.
A hand that holds an object carries it along and is never pushed downward, and in a pair with a free hand it moves at most 10\,mm.
A hand against the other arm is checked at recording (\cref{sec:supp_recording}).

\subsection{Retiming}
\label{sec:supp_retime}
Each interval $i$ between frames is stretched by a factor $\sigma_{i}$ set by its largest velocity ratio $\rho_{i}$ over all arm and finger joints,
\begin{equation}
\rho_{i} = \max_{j}\frac{\lvert \hat{q}_{j,i+1}-\hat{q}_{j,i}\rvert}{v_{j}\,\Delta t},\qquad
\sigma_{i} = \max(1,\;\gamma\, \rho_{i}),
\label{eq:retime}
\end{equation}
where $\hat{q}_{j,i}$ is joint $j$ at frame $i$, $\Delta t=0.02$\,s the frame interval and $\gamma=1.15$ a margin.
An interval is thus stretched only where its fastest joint exceeds $1/\gamma\approx 87\%$ of its limit.
Every channel is resampled linearly at 50\,Hz on the new time axis $t'_{k}=\sum_{i<k} \sigma_{i}\Delta t$, which lengthens the clip from $T$ to $T'$ frames.
The objects slow down with the robot.
The reference thus keeps their demonstrated path, apart from the hand-separation shifts of \cref{sec:supp_adjust}, but not its timing.
The \emph{retiming ratio} $T'/T$ is the mean of $\sigma_{i}$.
A candidate fails if its retiming ratio exceeds three, a bound that the arm IK and the arm refinement already apply before retiming.
A final check on the finished reference re-measures the Wrist and fingers row and the joint-speed bounds of the Timing row of \cref{tab:supp_filters}, which retiming meets by construction.
Over the clips that our search keeps in \cref{tab:abl_place}, the retiming ratio has a mean of 1.15 and a 90th percentile of 1.42.

%% file: sections/S3_supp_rl.tex
\section{Physics correction details}
\label{sec:supp_details}

\subsection{Bundles}
\label{sec:supp_bundles}
A bundle holds clips of one dataset.
They share the controlled arms, the active hands, whether the two hands share one object, the joint type of each articulated object that a hand manipulates and whether the scene has a table, since all parallel environments of a training run must hold the same kinds of bodies.
Each of the 4,096 parallel environments of a training run follows the reference of one clip of the bundle for the whole run, and every clip thus has at least 64 environments.
The policy takes no clip identifier as input and thus distinguishes the clips of a bundle only through their references and objects.

A two-arm bundle controls both arms, also in one-hand clips, where the policy holds the inactive hand and its arm at a rest pose under their hand-tracking terms.
This hand observes zeros in place of an object state and has no object or contact terms.
A one-arm bundle of one-hand clips controls only the active arm and holds the other arm and hand at a rest pose.

The clips that may share a training run are packed into as few bundles of at most 64 clips as possible, of nearly equal size, with objects of similar shape together.

\subsection{Residual action}
\label{sec:supp_action}
Physics runs at 100\,Hz, two steps per policy step.
A hand has $F=20$--$22$ finger joints, and its residual has $6+F$ components.
The residual $\mathbf{u}_{t}$ is the exponential moving average, with weight 0.4 on the new value, of the raw policy output $\tilde{\mathbf{u}}_{t}$, which is unbounded.
The bounds of \cref{eq:wrist} are $\delta_{p}=7$\,cm and $\delta_{R}=120^\circ$ per component of the position and of the rotation vector, and the look-ahead is $\ell=5$ frames (0.1\,s).
The position correction acts along the axes of a fixed frame at the table top, and the rotation correction in the wrist frame.
The policy changes the corrected wrist pose $(\mathbf{p}^{*}_{t},\mathbf{R}^{*}_{t})$ in steps at 50\,Hz, which a stiff arm would turn into torque spikes.
This pose is therefore low-pass filtered with weight 0.4 before IK.
Damped least-squares differential IK, with damping 0.05 and without a null-space term, converts it to the seven arm joint targets.
The finger targets are
\begin{equation}
\mathbf{q}^{*}_{f}=\mathbf{q}^{\mathrm{lo}}+\tfrac{1}{2}\big(\mathrm{sat}(\mathbf{b}_{f,t}+\beta\,\mathbf{u}^{f}_{t})+1\big)\odot(\mathbf{q}^{\mathrm{hi}}-\mathbf{q}^{\mathrm{lo}}),
\label{eq:finger}
\end{equation}
where $\mathbf{b}_{f,t}$ is the reference angle at the current frame, normalized to $[-1,1]$ by the joint limits $\mathbf{q}^{\mathrm{lo}},\mathbf{q}^{\mathrm{hi}}$, $\beta=1.5$, and $\mathrm{sat}$ is as in \cref{eq:wrist}.
The fingers have no look-ahead, since they lag their targets little and targets $\ell$ frames ahead would close them before the hand reaches the object.
The joint torques are
\begin{equation}
\boldsymbol{\tau}=K_{p}(\mathbf{q}^{*}-\mathbf{q})-K_{d}\dot{\mathbf{q}}+\boldsymbol{\tau}_{g},
\label{eq:pd}
\end{equation}
with $K_{p}=900$, $K_{d}=90$ and gravity compensation $\boldsymbol{\tau}_{g}$ for the arms, and the finger gains of \cref{sec:supp_gains}.
At constant speed, the PD loop of an arm lags its target by $K_{d}/K_{p}=0.1$\,s.
An arm reduces this lag with a velocity lead $(K_{d}/K_{p})\,\dot{\hat{\mathbf{q}}}_{t+\ell}$ added to its position target, with $\dot{\hat{\mathbf{q}}}$ a smoothed velocity of the reference arm motion.
The look-ahead $\ell$ compensates for the remaining lag of the arm controller, which includes the low-pass filter before IK.
The position targets $\mathbf{q}^{*}$ are the actions that \cref{sec:data} records.

\subsection{Observation and rewards}
\label{sec:supp_obs}
\input{sections/floats/tab_supp_rewards}
\noindent\textbf{Observation.}
Each hand's own state holds the wrist position, 6D rotation and linear and angular velocity in a fixed frame at the table top, the finger angles and velocities, and the gravity direction in the wrist frame.
All other geometric inputs are relative to the wrist, except the object rotation error, which is in the object frame.
The wrist and fingertip inputs hold the pose or position errors and velocity errors to the reference and the reference velocities.
The hand also observes its base action, the action at zero residual (zero at the wrist and $\mathbf{b}_{f,t}$ at the fingers), and its previous raw output $\tilde{\mathbf{u}}_{t-1}$.
The object state holds the pose and velocity of the manipulated object, its tracking errors and reference angular velocity, the vectors, normals and signed distances from the fingertips to its surface, the fingertip contact forces, its friction, restitution and mass, and, for an articulated object, the joint angle and its error.
The fingertip force is the net normal contact force from all contacts, not only the object.
The policy reads these contact forces and physical parameters from the simulator, in training and in data generation alike.

\noindent\textbf{Rewards.}
\Cref{tab:supp_rewards} lists the reward terms of each hand.
Since the reference fingertip positions come from retargeting a human hand of another shape, a fingertip earns its full tracking reward anywhere within 2\,cm of its reference position, which leaves room for a grasp that suits the robot hand.
Fingertip errors are measured in a fixed frame, not the wrist frame, and a wrist deviation thus also uses up this tolerance.
The contact terms (contact matching, grip force, grip progress, pre-grasp palm and force direction in \cref{tab:supp_rewards}) draw the fingertips toward the demonstrated contact points and reward grip force.

\subsection{Rollouts and termination}
\label{sec:supp_termination}
The termination window reaches from 0.5\,s behind to 0.1\,s ahead of the current frame.
Each manipulated object must match its reference pose at one of the window's frames within 5\,cm in position and 90$^\circ$ in rotation together.
An articulated object also ends a rollout when its joint error stays above 0.5\,rad, or half the range for a slide joint, for 0.5\,s.
A failure of either hand ends the rollout for both.
After a reset, the object-pose test is disarmed for 1.2\,s in two-arm bundles and 0.4\,s in one-arm bundles, in training and at evaluation.
Two-arm bundles also hold the objects at their reference for the first 0.2\,s, so that overlaps at a reset do not push them away.

Only the training of one-arm bundles also ends a rollout when no fingertip presses the object with at least 0.05\,N for 1.5\,s while the demonstration has contact.
Evaluation has no such test, since the test serves only to shape training and success is defined by the objects, as in \cref{eq:goal}.
Completion thus depends only on the objects, as the only other terminations detect simulation instability, such as non-finite states or speeds above any in the references.

At the end of training, acceptance evaluates each clip at the last two checkpoints, at 7/8 of training and at its end, with at least 64 rollouts per clip at each from its first frame, using the mean action without domain randomization.
These rollouts still differ, as small numerical differences grow under contact.
A clip is accepted if, at either checkpoint, at least 10\% of its rollouts complete.
An accepted clip is recorded with the checkpoint at which it completes most often (\cref{sec:supp_recording}).
The comparisons of \cref{tab:same64,tab:abl_rl} score only the final checkpoint.

\subsection{Training}
\label{sec:supp_training}
Training runs 1,600 iterations with 4,096 parallel environments.
PPO uses 32 steps per update, five epochs, 16 mini-batches, learning rate $5\times10^{-4}$, discount 0.99, GAE parameter 0.95, clip ratio 0.2, entropy coefficient 0.01, gradient-norm limit 1.0, early stopping of an epoch once the KL divergence summed over the two hands exceeds 0.04, and seed 42.
One-arm bundles use a KL threshold of 0.02 and eight mini-batches.
Each hand's actor is a transformer (two layers, width 64, four heads) over its own 17 input tokens, one per input group and two per fingertip, with a learned state-independent log standard deviation.
In two-arm bundles, the critic reads the tokens of both hands through one trunk, and each hand's advantage, from its own reward and value, updates only its actor.

Half of the training rollouts start at frame 0, and the other half, the \emph{random starts}, at random frames, so that later parts of a clip are learned early.
Their start frame is drawn uniformly from a range that grows from the first 35\% of the clip to the whole clip by 40\% of training, and the robot and objects are set to their reference poses there.
The termination thresholds start at 10\,cm and 180$^\circ$ and tighten to their final values over the first half of training, so that early rollouts last long enough to learn from.
Over the first 10\% of training, gravity on the manipulated objects ramps up from zero, so that a grasp can form before it has to bear the object's full weight, and their friction decays from a raised value to its final one.
The object-tracking weights of \cref{tab:supp_rewards}, the far keypoint and the articulation angle included, rise linearly from $1\times$ to $4\times$ over the first 75\% of training, so that grasped objects are lifted as demonstrated.

\subsection{Finger gains and effort limits}
\label{sec:supp_gains}
For Wuji~2, XHand Pro and Tesollo (DG-5F-S), the finger gains follow the effort limit of each joint, with $K_{p}$ reaching it at a 0.2\,rad error and $K_{d}=K_{p}/4$, so that typical tracking errors stay below saturation.
This limit is the value of the vendor model for Wuji~2 and XHand Pro and, for Tesollo, the stall torque in its manual.
Since the models of Wuji~2 and XHand Pro give less fingertip force than the vendors state, we raise their applied limits, but not their gains.
The fingertip force then reaches 12\,N for Wuji~2, its stated value, and about 9\,N for XHand Pro.
Tesollo starts at $3\times$ its limit, which falls back to the limit between 25\% and 50\% of training.
SharpA uses its specified effort values, with its own gains.
The XHand Pro velocity limits are up to $3\times$ their model values, which its vendor's stated closing rate allows.
Retiming (\cref{sec:supp_retime}) and the Timing row of \cref{tab:supp_filters} also use these raised XHand Pro limits.

\subsection{Scene in simulation}
\label{sec:supp_scene}
Manipulated objects are rigid or articulated dynamic bodies that the hands move through contact.
Each articulated object that a hand manipulates has one hinge or slide joint.
Every manipulated object weighs 0.2\,kg, an articulated one in total over its links, and has zero restitution.
Domain randomization scales the mass of a manipulated object by $\mathcal{U}(0.85,1.15)$, also randomizes its friction, and shifts its center of mass by $\pm1$\,cm.
Halving or doubling this mass at evaluation leaves MP-SR unchanged (\cref{tab:supp_robust}).

Manipulated objects collide with each other, with the table, with the robot including the torso, with their supports and with the fixed objects that receive them.
The other objects that the hands move, the \emph{other moved objects}, follow their reference poses kinematically.
They collide with the hands and arms, since a hand can carry them as well, but not with the manipulated objects, which a kinematic body would push regardless of the policy.
They have no reward or termination term, and the policy does not observe them.
The robot collides with the table and with the tracked supports, which do not collide with the table.
The added discs and the other fixed objects, the still objects that a hand touches or that receive another object, do not collide with the robot.
Still objects that nothing touches appear only in rendering.
Collisions between the two arms and between an arm and the torso, hands included, are off, so that they do not constrain the policy's exploration.
The arm IK and hand separation avoid most of these penetrations instead (\cref{sec:supp_armik,sec:supp_adjust}). 
A check after recording then excludes the few trajectories that still penetrate deeply (\cref{sec:supp_recording}).

%% file: sections/floats/tab_supp_rewards.tex
\begin{table*}[t]
\centering
{\footnotesize\setlength{\tabcolsep}{4pt}
\renewcommand{\arraystretch}{1.1}\begin{tabular}{@{}lrl@{}}
\toprule
Term & Weight & Form \\
\midrule
\multicolumn{3}{@{}l}{\textit{Hand tracking}} \\
Wrist position & 0.6 & $\exp(-40\lVert\Delta\mathbf{p}_{w}\rVert)$ \\
Wrist rotation & 0.6 & $\exp(-3\,\theta_{w})$ \\
Fingertips (thumb\ldots pinky) & 0.9, 0.8, 0.75, 0.6, 0.6 & $\exp(-k\max(0,\lVert\Delta\mathbf{p}_{\mathrm{tip}}\rVert-0.02))$, $k$ = 100, 90, 80, 60, 60 \\
Proximal links & 0.5 & $\exp(-50\,\overline{\lVert\Delta\mathbf{p}\rVert})$ \\
Wrist velocity & 0.1 & $\exp(-\lVert\Delta\mathbf{v}_{w}\rVert)$ \\
Finger joint velocity & 0.1 & $\exp(-\overline{\lvert\Delta\dot{q}_{f}\rvert})$ \\
\midrule
\multicolumn{3}{@{}l}{\textit{Object tracking} (weights at $1\times$; ramped to $4\times$)} \\
Position & 2.5 & $\exp(-80\lVert\Delta\mathbf{p}_{o}\rVert)$ \\
Rotation & 2.0 & $\exp(-3\,\theta_{o})$ \\
Far keypoint & 2.5 & $\exp(-80\lVert\Delta(\mathbf{p}_{o}+\mathbf{R}_{o}\boldsymbol{\kappa})\rVert)$ \\
Linear, angular velocity & 0.1, 0.1 & $\exp(-\overline{\lvert\Delta\mathbf{v}_{o}\rvert})$, $\exp(-\overline{\lvert\Delta\boldsymbol{\omega}_{o}\rvert})$ \\
Articulation angle & 2.0 & $\exp(-3\lvert\Delta\vartheta\rvert)$ \\
\midrule
\multicolumn{3}{@{}l}{\textit{Contact and force}} \\
Contact matching & 0.5 & $\sum_{f\in\mathcal{C}_{t}}\exp(-15\,d_{f})$ \\
Grip force & 10.0 & $\sum_{f}w_{f}\min(\phi_{f},5)/5\,/\sum_{f}w_{f}$ over touching fingers, $w$=1 thumb, 1.5 others \\
Grip progress & 2.0 & $\sum_{f}w_{f}\,\mathrm{clip}(d_{f,t-1}-d_{f,t},-0.01,0.01)/0.01\,/\sum_{f}w_{f}$ over approaching fingers \\
Pre-grasp palm & 2.0 & $\exp(-8\max(0,\lVert\mathbf{p}_{\mathrm{palm}}-\bar{\mathbf{p}}_{c}\rVert-0.12))$ \\
Force direction & 2.5 & $\mathrm{clip}(0.5+0.5\,\phi_{h}/30,0,1)$, scaled by the share of thumb and index in contact \\
\midrule
\multicolumn{3}{@{}l}{\textit{Regularizers}} \\
Finger residual & $0.9\cdot6/F$ & $-\max(0,\lVert\tilde{\mathbf{u}}_{f}\rVert-0.5\sqrt{F/6})^{2}$ \\
Wrist residual & 0.9 & $-\sum_{i}\max(0,\lvert\tilde{u}_{i}\rvert-\epsilon)^{2}$, $\epsilon$: $0\to0.7$ \\
Action rate & 0.1 & $\exp(-\lVert\tilde{\mathbf{u}}_{t}-\tilde{\mathbf{u}}_{t-1}\rVert_{1})$ \\
Wrist, finger action rate & 0.3, $0.1\cdot6/F$ & $-\lVert\tilde{\mathbf{u}}_{t}-\tilde{\mathbf{u}}_{t-1}\rVert^{2}$ over the wrist or finger part \\
\bottomrule
\end{tabular}}
\par\smallskip
{\footnotesize\raggedright Symbols. $F$: number of finger joints; $g$: gravity; $\Delta$: difference from the reference; an overline: mean over the elements; $\theta$: rotation angle between current and reference orientation; $\boldsymbol{\kappa}$: object-frame position of the surface keypoint farthest from the object's centroid; $\Delta\vartheta$: joint error, in rad for a hinge and as a fraction of the range for a slide; $d_{f}$: distance from fingertip $f$ to its human contact point, the median in object coordinates of where that finger touched the manipulated object in the demonstration; $\mathcal{C}_{t}$: fingers in contact in the demonstration at frame $t$; touching fingers: those that touch the manipulated object somewhere in the demonstration and now press it with at least 0.05\,N, from the clip start until two (thumb) or five frames before their last demonstrated contact; approaching fingers: the same fingers while they press with less than 0.05\,N, in the same window but not before the hand's first demonstrated contact; $\phi_{f}$: fingertip--object force; $\phi_{h}$: force of the hand on the manipulated object along $\hat{\mathbf{a}}_{o}+g\hat{\mathbf{z}}$, the direction of the force that the reference motion requires, counted while the reference object moves, from three frames before the first demonstrated contact; $\hat{\mathbf{a}}_{o}$: reference object acceleration; $\bar{\mathbf{p}}_{c}$: mean of the human contact points; $\epsilon$: penalty-free band of the wrist residual, which widens between 10\% and 25\% of training; $\tilde{\mathbf{u}}$: raw policy output; proximal links: the finger links next to the palm.\par
}
\caption{Reward terms per hand, with lengths in m, angles in rad and forces in N. The symbols are defined above.}
\label{tab:supp_rewards}
\end{table*}

%% file: sections/S4_supp_generation.tex
\section{Train-ready data generation details}
\label{sec:supp_generation}
\input{sections/floats/tab_spec}
\Cref{tab:spec} lists the data contents and their approximate size, and \cref{fig:supp_gallery_episode} shows them for one episode.

\subsection{Recording}
\label{sec:supp_recording}
Recording runs 64 rollouts per clip, then 128 and 256 if none completes, and a clip with no completed rollout is not recorded.
The kept rollout has the lowest mean, over its manipulated objects, of the position and rotation errors at the last frame, each divided by its termination threshold.
Since completion already keeps the objects near their reference through the termination test, this error at the last frame only ranks rollouts that all track the objects.

The recorded contact forces are the normal forces on each hand link, in total and per object.
Recording uses the object parameters of \cref{sec:supp_scene} without randomization.
The data also hold each manipulated object's reference pose and tracking error, the joint angles of articulated objects and the policy's raw output.
Each trajectory's states and actions come from one rollout.

Since these collisions are off in simulation (\cref{sec:supp_scene}), recording checks every trajectory again for penetrations between the two arms and between an arm and the torso, hands included, on the collision shapes of the simulation.
Our data exclude the trajectories in which an arm or its hand penetrates the torso or a link of the other arm by more than 2\,cm (\cref{tab:supp_selfpen}).
This limit is looser than the 2\,mm of the reference (\cref{tab:supp_filters}), which most recorded trajectories still meet (\cref{fig:supp_selfpen}).

\input{sections/floats/fig_supp_selfpen}

\subsection{Rendering and appearances}
\label{sec:supp_rendering}
Rendering runs no physics and sets, for each 30\,fps image, the recorded 50\,Hz state nearest in time.
The egocentric camera uses the rectified intrinsics of the real ZED Mini and its mounting pose calibrated with EasyHeC~\cite{easyhec}, and depth is the distance to the image plane of the left eye.
The head pitch is set once per clip, and all its trajectories keep it.
It centers the table in the vertical field of view, or puts the table's near edge at the bottom of the image when the table spans more than this field.
Without a table, or when too few of the hands and objects would be in view, it aims at them instead.
The stored head joint states and targets take this pitch.
The exocentric camera is the frontal scene camera, moved back until the robot's upper body fits in the image.

Simulation realizes the placement's height $z$ by moving the whole robot vertically, with its torso joints fixed.
Rendering puts the RB-Y1's wheeled base back on the floor by re-solving its torso pitch joints, which then stay fixed over the motion, so that the torso link, which carries the arms and defines the robot base frame, keeps its pose.
Where that fails, rendering shifts the rest of the scene instead and stretches or shortens the table legs.
A scene without a table is always shifted.

Of a trajectory's five appearances, one is fixed and four are random.
A random appearance varies the table, floor and wall materials, the lighting and the tone-mapping saturation.
Each trajectory adds up to three distractor objects on the table, the same in its four random appearances.
About half of a clip's random appearances add an indoor HDRI dome.
Distractors avoid the robot's and objects' paths and the cameras' lines of sight to the objects and hands.
Since ARCTIC has no table, its random appearances vary neither the table nor distractors.
Random appearances use assets from Poly Haven, ambientCG, Google Scanned Objects~\cite{gso} and NVIDIA Isaac.
Per-clip seeds make every appearance reproducible.

An episode leaves out the rendering-only still objects (\cref{sec:supp_scene}) that, for example, the robot would touch, that would hide a moving object or that would hang over the table edge.
Each episode stores, with the floor at zero height, the robot's base position and torso joints, the table's position and size, the added discs, the camera poses and how much of the active hands and objects the left egocentric image shows, so that episodes with hidden objects can be found.

\subsection{Language}
\label{sec:supp_language}
The model, Claude Sonnet 5.5, receives frames at the phases from an episode in the fixed appearance and a summary with the source task label where one exists, the objects with their roles, the active hands, the phases with their times and the object motion.
Each clip gets an instruction with four paraphrases, a short and a detailed description, and for each phase a sub-instruction with two paraphrases.

A hand touches an object when the contact force between one of its links and the object exceeds 0.5\,N, and an object moves once it is 1\,cm or 10$^\circ$ from the pose it holds just after the start.
Approach lasts until the first contact, grasp until an object moves, manipulate until the objects stop, release until the last contact and retreat to the end, and very short phases merge into a neighbor.
Each episode stores the phase spans of its trajectory with the sub-instructions of the clip.

Rule checks reject text that names the robot or the human, or that gives table- or viewer-relative directions such as to the left, forward or toward you, since one text serves every robot hand, placement and appearance of a clip.
They also test the text against the summary, for example which hand acts and whether an object turns far enough to be called flipped.
Claude Opus 5.5 generates again any text that fails these checks or that Claude Sonnet 5.5 rates as low in confidence.
A clip whose text still fails keeps its source label as its task text.

\subsection{Pose augmentation by re-simulation}
\label{sec:supp_augmentation}
A \emph{variant} is a placement other than the adopted one at the adopted height $z$.
We draw 32 variants per clip at random among the screening survivors at that height (\cref{sec:reference}), and 64 more at a time while fewer than four of them pass the arm IK.
A variant moves the objects and the active hands' reference rigidly over the table and thus keeps the retargeted hand motion with its tilts and the initial-pose repair with its lead-in.
For each variant, we rerun the arm IK and the retiming, and also the hand separation if the adopted placement needed it.
A variant whose arm IK fails is dropped.
The residual policy then runs at each variant from the checkpoint with which the clip was recorded, without retraining.

A variant's reference passes a reduced set of the checks in \cref{tab:supp_filters}.
It keeps the 2\,mm penetration tests of the Collision row, the table-area test of the Objects row and the bound of 3 on the retiming ratio, for each arm and its fingers.
Retiming keeps its joint speeds within their limits at every frame, as for every reference (\cref{sec:supp_retime}).
It skips the Wrist and fingers row and the first-frame tests of the Collision row.

The passing variants are grouped into \emph{tiers} by how far they move or turn the objects, largest change first, and keep the random sampling order within a tier.
The first 16 in this order get eight rollouts each.
A variant's three best completed rollouts, ranked as in recording, are tested, and a rollout counts if its executed joints pass the penetration tests of \cref{tab:supp_filters}.
A rollout passes each of these tests if it stays within the limit of \cref{tab:supp_filters} or within 1\,mm of the deepest penetration among the adopted placement's three best completed rollouts, which run alongside the variants.
A variant is thus no worse than the adopted placement.
If fewer than four of the first 16 yield a counted rollout, the rest get eight rollouts too.
Each variant keeps its best counted rollout.

A greedy farthest-point order then ranks these variants tier by tier, from the adopted placement, in position and yaw normalized by the searched region.
We keep up to three of the first eight in this order, visible ones first.
A variant is visible if the bounding box of every manipulated object stays inside the left egocentric image, with a 5\% margin, on at least 90\% of the steps in the second after the objects first move.
A clip whose adopted placement is out of view keeps up to four.
The adopted placement is out of view if fewer than 90\% of the active-hand and manipulated-object points within 0.5\,s before or after the first object motion are in that image, even after the head pitch is re-aimed at them.
Such a clip also samples placements ranked by predicted visibility and tries them first.
Its variants replace the adopted trajectory unless the clip has none.

For the Wuji~2 hand, the residual policies ran at 31,272 variants, and 85.2\% of them yielded a counted rollout.
Of the accepted Wuji~2 clips, 78.0\% gained at least one variant trajectory and 65.8\% at least three.

The duration in \cref{tab:spec} assumes that a variant lasts as long as the clip's reference.
The table's mean lengths exclude the lead-in, while the hours of its clips, trajectories and episodes include it.

%% file: sections/floats/tab_spec.tex
\begin{table}[t]
\centering
{\footnotesize
\input{figs_tabs/tables/tab_spec_body}}
\caption{Contents of the data and their approximate size. Privileged signals are simulator ground truth.}
\label{tab:spec}
\end{table}

%% file: figs_tabs/tables/tab_spec_body.tex
\begin{tabular}{@{}l>{\raggedright\arraybackslash}p{0.75\linewidth}@{}}
\toprule
\multicolumn{2}{@{}l}{\textit{Amount}} \\
\hspace*{6pt}Hands & 4 (\cref{fig:funnel}) \\
\hspace*{6pt}Clips & 20,768 (5,048--5,361 per hand), 37 h \\
\hspace*{6pt}Trajectories & $\times$ $\approx$ 3 placements $\rightarrow$ $\approx$ 65,000, 116 h \\
\hspace*{6pt}Episodes & $\times$ 5 appearances $\rightarrow$ $\approx$ 325,000, 578 h \\
\midrule
\multicolumn{2}{@{}l}{\textit{Content}} \\
\hspace*{6pt}Ego RGB-D & stereo 1280$\times$720 + left depth, 30 Hz \\
\hspace*{6pt}Exo RGB & 1920$\times$1920, 30 Hz \\
\hspace*{6pt}Joint states & head, arms, hands, 50 Hz \\
\hspace*{6pt}Actions & absolute joint PD targets, 50 Hz \\
\hspace*{6pt}Privileged & object poses, per-link contact forces, 50 Hz \\
\hspace*{6pt}Language & instruction and phase sub-instructions per clip \\
\midrule
\multicolumn{2}{@{}l}{\textit{Metadata}} \\
\hspace*{6pt}Length & 5.7 s human $\rightarrow$ 6.3 s robot (mean) \\
\hspace*{6pt}Scene & table height, robot base and torso pose \\
\hspace*{6pt}Camera & stereo model of the ZED Mini \\
\hspace*{6pt}Format & LeRobot v2.1~\cite{lerobot} \\
\bottomrule
\end{tabular}

%% file: sections/floats/fig_supp_selfpen.tex
\begin{figure}[t]
\centering
\includegraphics[width=\linewidth]{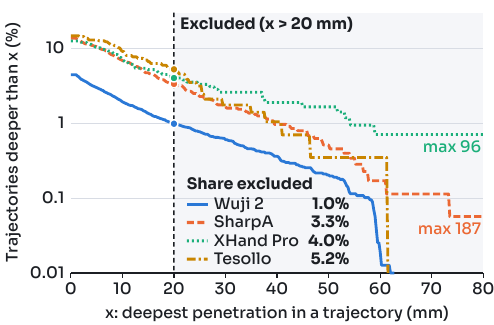}
\caption{Per hand, recorded trajectories whose deepest penetration of an arm or its hand into the torso or the other arm exceeds $x$. Legend: the excluded share, beyond 2\,cm (\cref{tab:supp_selfpen}). Max: deepest (mm) of a curve leaving the axis.}
\label{fig:supp_selfpen}
\end{figure}

%% file: sections/S5_supp_experiments.tex
\section{Experiment details}
\label{sec:supp_experiments}

\subsection{Statistics of the recorded data}
\label{sec:supp_stats}
\Cref{tab:supp_funnel} counts, for each robot hand and source dataset, the clips that pass each step of \cref{fig:funnel}.
Recording yields 15,796 Wuji~2 trajectories, 3.13 per accepted clip, with five episodes each.
Scored with the metrics of \cref{tab:same64} against their reference object motion, these trajectories reach a mean ADD-AUC of 0.935.
Of them, 94.2\% meet the object criteria of ManipTrans and 95.7\% those of SPIDER, and every source dataset reaches an ADD-AUC of at least 0.90.
These values describe only the completed rollouts that recording keeps (\cref{sec:supp_recording}) and measure how closely they follow their reference object motion, not how often RL succeeds.
The recorded data of the other three hands track as closely.
On a sample of up to 300 trajectories per hand and source dataset, SharpA, XHand Pro and Tesollo reach ADD-AUC 0.942, 0.937 and 0.933, and the same sampling gives 0.936 for Wuji~2.
Their trajectories meet the criteria of ManipTrans in 92.9 to 94.9\% and those of SPIDER in 94.8 to 96.4\%.

\input{sections/floats/tab_supp_funnel}
ARCTIC clips pass physics correction least often (\cref{tab:supp_funnel}).
Four in five of the ARCTIC clips that the arms can follow turn an articulated part with both hands and without a table.
Of these, 14 to 26\% pass physics correction, against 46 to 62\% of its two-hand grasps of a whole object.

\subsection{Conditions of the pipeline comparison}
\label{sec:supp_cmp_conditions}
DexMan~\cite{dexman}, OATH~\cite{oath} and GNR~\cite{gnr} have no released code, and \cref{tab:same64} does not include them.

\noindent\textbf{Clips.}
The pool holds the 520 of CHORD's released two-handed TACO clips that pass data preparation and kinematic reference generation, in 81 combinations of verb, tool and target.
With seed 42 we draw 64 combinations and one clip from each without using any method's ADD-AUC or success rates.
The draw skips clips whose motion file CHORD's loader cannot read.

\noindent\textbf{Conditions.}
ManipTrans, DexMachina, SPIDER, EgoEngine and FlashDexRetarget run with our object mass.
DexMachina, SPIDER and EgoEngine also use our hand--object friction and SharpA finger gains, force limits and armature.
CHORD runs as FlashCHORD v0.4, the version of its RL tracker in the code its authors released, and keeps its released assets, including their object mass and inertia.
FlashDexRetarget also runs with its support discs switched on, and it trains for 500M environment steps, the end of its default schedule.
SPIDER and EgoEngine run in scenes with our table plane, supports (\cref{sec:supp_layout}) and object collision meshes.
EgoEngine runs from SPIDER's inputs on the SharpA hand and controls both hands, unlike its TACO setting.
EgoEngine runs three times per clip, and we report the mean score over its runs and the time of one run.

\noindent\textbf{Cost and training.}
Costs are GPU-minutes per clip on an NVIDIA L40S.
ManipTrans and DexMachina train each clip until its learning curve flattens, for at most 150 minutes.
Trained for 200 iterations instead of 1,600, our single-clip policies reach ADD-AUC 0.940, SP-SR 98.4 and MP-SR 85.9 at 43.1 GPU-minutes per clip.
This is about a third of the GPU time of ManipTrans, DexMachina or CHORD in \cref{tab:same64}.
The reward-gated decay of the virtual object controllers of DexMachina does not start within its budget.
These controllers thus decay linearly to zero over the first two thirds of its training, and we score it after they reach zero.
Its objects then mostly stay within the 10\,cm of SP-SR but rarely within the 3\,cm of MP-SR.
Our floating hand is driven at the wrist by six virtual joints, three prismatic and three revolute.

\noindent\textbf{Scoring.}
Unlike acceptance, every rollout runs to the end of the clip with object-tracking termination off, and we score the final checkpoint.
Each step of a method is scored against the original object motion at the time that step maps to, so that a method that slows the motion down is compared with the same demonstration.
SP-SR requires a mean object error within 10\,cm and 0.5\,rad, the rule of SPIDER, and MP-SR requires the object criteria of ManipTrans, within 3\,cm and 30$^\circ$.
As SPIDER and EgoEngine output one trajectory per run, \cref{tab:supp_rollouts} also scores the same evaluations by the mean over rollouts, without a rerun.
Scored this way, the arm-mounted policies with $K=64$ still score above every prior method in every metric.

\input{sections/floats/tab_supp_rollouts}

\subsection{FlashDexRetarget and our recipe}
\label{sec:supp_flashdex}
{\looseness=-1
FlashDexRetarget~\cite{flashdexretarget} gives each hand its own actor and critic, while our two actors share one critic trunk with a value head per hand.
In its released code, the wrist adds a bounded residual to the reference as ours does, while the finger joints take absolute targets within their limits instead of shifting the reference angles.
It also observes ten future reference frames and object point clouds, rewards distances between hand and object instead of matching contact points, and trains off-policy with FlashSAC, a replay buffer of 50M transitions and a critic of width 1,024.
Our floating-hand policy trains for about 210M environment steps against FlashDexRetarget's 500M.
Of its 16-fold cost per clip in \cref{tab:same64}, a factor of 2.4 thus comes from more steps and a factor of 6.7 from a higher cost per step.\par}

\subsection{Arm IK after a floating-hand policy}
\label{sec:supp_posthoc}
\input{sections/floats/tab_supp_posthoc}
The wrist targets are the wrist poses that our floating-hand policy in \cref{tab:same64} reaches in its best rollout of each clip, with the fingers at their achieved positions and the placement that our search adopted for each clip, which favors this baseline.
The arm IK of \cref{sec:reference} runs unchanged on them and on the wrist targets of our reference, with the same checks, and keeps 26\% fewer clips on the floating-hand wrists than on those of our reference (\cref{tab:supp_posthoc}).
Most of the clips lost on the floating-hand wrists penetrate the table or the torso.
After our retiming, every motion kept on the floating-hand wrists stays within the joint position and velocity limits.
The last row of \cref{tab:supp_posthoc} runs the same arm IK, with the fingers of our reference, on the wrist trajectories of SPIDER's rollouts at our object mass.
It keeps half of the clips, counted before retiming.

\subsection{Placement ablation}
\label{sec:supp_abl_place}
\input{sections/floats/tab_abl_place}We draw 128 of the Wuji~2 clips left after data preparation with seed 42, in the proportions of the source datasets, and generate the reference of each under every row of \cref{tab:abl_place}.
Each row only chooses the placements, and the arm IK, retiming and the checks of \cref{tab:supp_filters} decide whether a clip is kept.
The default-position row tests only the zero shift and turn at the default scene position of \cref{sec:supp_candidates}, at every height $z$ of its dataset.
The 12 random candidates are drawn uniformly from the grid of \cref{sec:supp_candidates} with seed 42, and the first of them that passes is adopted.
The top screened candidate is the survivor with the most covered frames, tested alone.
Random order shuffles, with seed 42, every candidate of the grid that fits on the table, without the screening.
It stops as ours does (\cref{sec:supp_order}), and its second row has no limit on candidates tested.
The every-candidate row tests these candidates nearest first by the distance of the wrists from a natural arm pose with the elbow at 90$^\circ$, until three pass or 7,200 seconds elapse.
Unlike ours, it skips the screening and its coverage order, has no first round (\cref{sec:supp_order}) and sets no limit on candidates tested.

Over all 12,951 Wuji~2 clips that it tested, our search keeps 51.6\% (\cref{tab:supp_drops}), more than in this sample.
Of the difference of 19,629 seconds between our search and every candidate over the 128 clips, 78\% comes from the 44 clips that ours drops at its limit on candidates tested and 19\% from the 23 clips whose screened candidates run out.

\subsection{RL-design ablation}
\label{sec:supp_abl_rl}
The ablation trains the 64 clips of \cref{tab:same64} as one bundle on the arm-mounted SharpA hand, with seed 42 and 1,600 iterations.
Every run is scored at iteration 1,600 with the metrics of \cref{tab:same64}, but with early termination on.
A rollout thus ends once a manipulated object strays from every reference pose in the termination window (\cref{sec:supp_termination}), and its missing steps count as failures.
No change in \cref{tab:abl_rl} alters the number of environments or the network width.
In the rows that re-solve the arm by IK at random starts, every reset puts the arms at joints that IK solves for the wrist pose of that frame alone, from a neutral configuration and seeded random ones, within 2\,mm and 1$^\circ$.
This holds at frame 0 and at random frames, in training and evaluation alike.
The reference that the policy observes and follows stays ours, as do the arms' gravity compensation and the velocity lead, which comes from our arm motion.
The shares in \cref{sec:exp_data} of frame pairs in which some arm joint moves by more than 20$^\circ$ are over the 18,403 pairs of adjacent frames per side in the 64 clips.

With residuals on the arm joints, the policy adds one bounded residual per arm joint to the reference arm joints and sends the sum as the arm joint targets without IK, and it also observes the errors of the arm joints to the reference and their velocities.
Turning off the contact terms removes contact matching, grip force, grip progress, pre-grasp palm and force direction (\cref{tab:supp_rewards}).
Strict termination ends a rollout at the first frame whose reference pose the object misses, without the termination window or the disarmed period after a reset.
Without the low-pass filter, the corrected wrist pose goes to IK unfiltered.
\Cref{tab:supp_ablation} lists each change with all metrics.
Removing the contact terms gives the largest drop among the reward terms, from 0.900 to 0.837 in best ADD-AUC and from 81.2 to 70.3 in MP-SR.
Block (d) of \cref{tab:supp_rl_more} repeats the full recipe three times, and its best ADD-AUC spans 0.896 to 0.900.
Training without random starts, also repeated three times, spans 0.883 to 0.892.
The two ranges do not overlap in SP-SR or MP-SR either.
Across all changes repeated in block (d), one change's best ADD-AUC differs by up to 0.019 between runs, and its SP-SR by up to 9.4 points.
A single run that differs from the full recipe by less than this does not separate from it.

A physics test evaluates the $K=64$ arm-mounted policies of \cref{tab:same64} again, without retraining, on the same 64 clips and with the same scoring (\cref{tab:supp_robust}).
It halves or doubles the mass of every manipulated object, outside the $\pm$15\% of domain randomization, and the mass the policies observe changes with it.
It also halves the friction of the manipulated objects and of the robot, which halves the friction of every hand--object contact, beyond the friction randomization of training.
Neither mass changes MP-SR, and SP-SR drops by at most 1.5 and ADD-AUC by at most 0.018.
Halved friction lowers MP-SR from 84.4 to 73.4 and SP-SR from 98.4 to 95.3.

\input{sections/floats/tab_supp_ablation}
\input{sections/floats/tab_supp_robust}

\subsection{Arm solution ablation}
\label{sec:supp_abl_arm}
\noindent\textbf{Sequential IK.}
On the same 64 clips of \cref{tab:same64}, sequential IK solves each frame of the reference's wrist poses at the adopted placement from the previous frame's solution, and the first frame from the robot's rest pose.
It is held to the limits of our arm IK (\cref{sec:supp_armik}) and to the joint velocity limits.
Retimed by our rule, it keeps 28\% of the clips within every limit (the single run of \cref{sec:exp_data}), with motions a median 14\% longer than our retimed reference.
If the first frame is solved in one step from the rest pose, instead of being reached over 100 interpolated wrist poses, the run keeps 1 of the 64 clips within the limits other than velocity, and its tracking error exceeds the limit in 63.
Trained on all 64 of these retimed references with the setup and scoring of \cref{tab:abl_rl}, our RL recipe reaches ADD-AUC 0.874, SP-SR 73.4 and MP-SR 68.8, below the three runs on our reference (\cref{sec:supp_abl_rl}).
Trained on the sequential-IK references with our timing instead of their own retiming, our recipe reaches ADD-AUC 0.885 (\cref{tab:supp_rl_more}, block c).
A single sequential run can also start from the first-frame solution of our arm IK instead of the rest pose.
Trained on its references at our timing, our recipe reaches ADD-AUC 0.907 (\cref{tab:supp_rl_more}, block c).
The branch chosen at the first frame thus decides the result, and our arm solution chooses this branch with the whole clip in view (\cref{sec:reference}).

\noindent\textbf{Parts of our arm solution.}
Removing parts of our arm solution on the same 64 clips, under the arm IK's own checks before retiming, keeps 59\% of the clips without the choice among runs, 83\% without re-solving, 70\% without the coarse path and 9\% without all three, against 89\% for the full solution.
Unlike the sequential IK above, which keeps 28\% after retiming, the variant without all three starts its one run from a random configuration instead of the rest pose.
Removing only the re-solving from the coarse path still keeps 89\%.
The gain of the coarse path, from 70\% to 89\% of the clips, thus comes from the run that starts from it.

\subsection{Scenes for policies in simulation}
\label{sec:supp_policy_scenes}
The 20 clips of each task are accepted Wuji~2 clips, at most one per object.
The adopted placement of each passes the task's success rule, also with added noise, and shows the object to the head camera at the grasp.
A scene is one clip at one placement, and it runs once, in a fresh process with one rollout.
For these tasks, the pose augmentation of \cref{sec:supp_augmentation} tests the placements in a box of $\pm$10.5\,cm and $\pm$30$^\circ$ around the adopted one, nearest first rather than far apart, and each clip keeps up to 100 variants that pass the pipeline.
Five of them, chosen with seed 42, are held out for L1, and the rest train, each rendered with up to 20 visual augmentations (\cref{sec:data}), at 320$\times$180 and 30\,Hz.
L0 and L1 hold 100 scenes per task, five per clip, that also pass a replay of the recorded actions and two replays with added noise.
Where too few held-out placements passed, placements for L1 came from a box widened to at most 17.5\,cm and 35$^\circ$, which training does not use.
Ours, kinematic replay and SPIDER + arm use the same scenes.
The SPIDER + arm data retarget the same 20 clips per task with SPIDER and attach our arms by SPIDER's own rule.
They share the placements, appearances and episode count of the kinematic replay data.
In evaluation, a SPIDER + arm scene starts from the first state of SPIDER's trajectory.
In addition to the success rules of \cref{sec:supp_success}, a rollout fails if an object exceeds 5\,m/s or moves at least 2\,cm in the first 0.1\,s.

\subsection{Policies in simulation}
\label{sec:supp_policy_training}
The policy observes the 56-dimensional joint state and the head stereo pair, and its action is the absolute joint target of the same joints, without the torso joints.
Diffusion Policy observes two frames, predicts 16 actions and executes eight, and trains for 166,650 steps at batch size 512 and learning rate $2.83\times10^{-4}$.
GR00T N1.7 has an action horizon of 40 frames and a frozen visual encoder, and trains at batch size 256 and learning rate $1.5\times10^{-4}$, for 40,000 steps on our data and on SPIDER + arm data, and for about 150 passes over its data on kinematic replay (\cref{tab:supp_e5_long}).
The mid-training data are the recorded Wuji~2 data without the source sequences of the task clips, 73,665 episodes with 13.5 million frames.
Mid-training starts from the public weights and runs 100,000 steps at batch size 512 over them.
GR00T + mid and GR00T on our data then fine-tune with one recipe for 40,000 steps at batch size 256, from the mid-trained and the public weights.
Every policy is one run with seed 42, and we evaluate its final checkpoint under the appearance of its training data.

\Cref{tab:supp_e5_long} adds GR00T and GR00T + mid fine-tuned on our data for about 150 passes, the length used on kinematic replay.
At that length GR00T reaches 82\% at L0 and 50\% at L1 without mid-training and 81\% and 55\% with it, against 6\% and 5\% on kinematic replay.

\input{sections/floats/tab_supp_e5_long}
Success spreads over the clips of a task.
With our data, Diffusion Policy succeeds in at least one scene of 15 to 20 of the 20 clips per task and level, and GR00T fine-tuned for about 150 passes in 16 to 20.
With kinematic replay data, Diffusion Policy and GR00T at about 150 passes reach at most 2 and 4 such clips.
With SPIDER + arm data, Diffusion Policy and GR00T, the latter fine-tuned for 40,000 steps, reach at most 3 such clips each.

\subsection{Success rules in simulation}
\label{sec:supp_success}
A scene is evaluated only if the reference passes its rule and the object held still at its first frame fails it.
A P\&P rollout must lift the object, which requires that it rises by at least half of the maximum rise in the scene's recorded trajectory, stays above that trajectory's lowest height minus 5\,cm and stays lifted for at least 0.3\,s while a hand holds it with more than 1\,N.
At the last frame, the object's mesh center must lie within 5\,cm of its position in the last frame of the kinematic reference, and the object must not be turned over by more than 135$^\circ$ from it.
Pour requires the tool to stay within 5\,cm and 45$^\circ$ of its reference pose relative to the receiving object for 0.2\,s, at the frame of largest tilt in the scene's kinematic reference, while the receiving object tips by at most 45$^\circ$.
Sweep requires the tool to travel in contact with the target while a hand holds it with more than 1\,N, for 10\,cm or half of the travel in the scene's recorded trajectory, whichever is smaller.

\subsection{Real-robot experiments}
\label{sec:supp_real}
\input{sections/floats/fig_supp_realaug}
Most objects are 3D-printed in green from the dataset meshes, and the brush is a real object.
The clips of \cref{tab:real} are the alarm clock (P\&P), bowl to cup (Pour) and brush to box (Sweep).
They are task clips of \cref{sec:supp_policy_scenes}, whose source sequences the mid-training data leave out.
GR00T and GR00T + mid fine-tune from the public and the mid-trained weights for the same 40,000 steps, with state dropout 0.3.
Every training episode starts from one fixed pose per clip, with the fingers relaxed.
The fine-tuning episodes show each placement variant under 35 appearances (\cref{fig:supp_realaug}), 12 of them with the manipulated objects in the green of the prints.

Replay sends the recorded joint trajectory of the clip at the tested placement open loop, without a policy.
Each trial is scored from its video, and a P\&P trial succeeds if the object is lifted and ends on the pad that marks its target, a Pour trial if the tool reaches the pouring pose over the receiving object, and a Sweep trial if the held tool moves along the target in contact.
L0 tests five placements used in fine-tuning and L1 the five placements left out, one trial each.

\subsection{End of the placement search}
\label{sec:supp_drops}
\input{sections/floats/tab_supp_drops}
Testing every candidate keeps only two more of the 128 sampled Wuji~2 clips than our search (\cref{tab:abl_place}), which suggests that few of the dropped clips have a placement.
The last row of \cref{tab:supp_drops} counts clips that fail a check after the search, in the adjustments, retiming and final check of the reference (\cref{sec:supp_adjust,sec:supp_retime}) or in building the robot scene.

\subsection{Further RL ablations}
\label{sec:supp_rl_more}
\input{sections/floats/tab_supp_selfpen}
\input{sections/floats/tab_supp_rl_more}
\Cref{tab:supp_rl_more} extends the RL-design ablation of \cref{sec:supp_abl_rl}, with its setup and scoring.
In (a), each row adds components to the row above, from plain tracking to the full recipe.
The second row adds the arms' gravity compensation and velocity lead, the termination window with its disarmed period, the curriculum of the termination thresholds, a reward and value head per hand, the action-rate penalties, the per-element wrist-residual free band, domain randomization and the reference velocity of objects at the start frame.
Plain tracking lacks these and every component of the later rows.
The second row alone lowers best ADD-AUC from 0.748 to 0.590.
Random starts raise it to 0.629 with the arm re-solved by IK, and to 0.789 with the arm reset to our reference, the first row above plain tracking.
These rows do not yet have the contact terms.
In (b), the floating-hand recipe of \cref{tab:same64} trains with one change at a time.
Its plain tracking is that of (a) without the arm.
Scored with early termination on, unlike \cref{tab:same64}, its rows compare only with each other.
On the floating hand, best ADD-AUC stays at 0.891 without random starts and falls to 0.853 without the contact terms.
Block (c) tests further changes, one run each.
Apart from the change that re-solves the arm at random starts, each lies within 0.03 of the full recipe in best ADD-AUC, while the same training repeated in block (d) spreads by up to 0.02.
None of these changes alters how the arm is attached, the subject of \cref{tab:abl_rl}.
The change that re-solves the arm at random starts, also without the look-ahead, gravity compensation, velocity lead and low-pass filter, scores lowest (0.863), as re-solving does in \cref{tab:abl_rl}.
Block (d) lists changes trained more than once, and \cref{sec:supp_abl_rl} compares the spread of its runs.

%% file: sections/floats/tab_supp_funnel.tex
\begin{table*}[t]
\centering
{\small\setlength{\tabcolsep}{2pt}
\input{figs_tabs/tables/tab_funnel}}
\caption{Clips per source dataset that pass each step, for each robot hand. Rows 1--3 count the clips that pass data preparation, kinematic reference generation and physics correction (row 1: mean over the hands; bold: final counts).}
\label{tab:supp_funnel}
\end{table*}

%% file: figs_tabs/tables/tab_funnel.tex
\begin{tabular}{@{}lrrrrrrrrrr@{}}
\toprule
 & \multicolumn{1}{c}{\rlap{\rotatebox{60}{HOT3D}}} & \multicolumn{1}{c}{\rlap{\rotatebox{60}{OakInk2}}} & \multicolumn{1}{c}{\rlap{\rotatebox{60}{TACO}}} & \multicolumn{1}{c}{\rlap{\rotatebox{60}{GRAB}}} & \multicolumn{1}{c}{\rlap{\rotatebox{60}{ARCTIC}}} & \multicolumn{1}{c}{\rlap{\rotatebox{60}{HRDexDB}}} & \multicolumn{1}{c}{\rlap{\rotatebox{60}{H2O}}} & \multicolumn{1}{c}{\rlap{\rotatebox{60}{HO-Cap}}} & \multicolumn{1}{c}{\rlap{\rotatebox{60}{DexCanvas}}} &  \\[-1pt]
 & \multicolumn{1}{c}{\chip{2A78D6}} & \multicolumn{1}{c}{\chip{EB6834}} & \multicolumn{1}{c}{\chip{1BAF7A}} & \multicolumn{1}{c}{\chip{EDA100}} & \multicolumn{1}{c}{\chip{E87BA4}} & \multicolumn{1}{c}{\chip{008300}} & \multicolumn{1}{c}{\chip{6250D6}} & \multicolumn{1}{c}{\chip{E34948}} & \multicolumn{1}{c}{\chip{7B8494}} & Total \\
\midrule
Sequences & 294 & 2,828 & 2,308 & 1,335 & 301 & 436 & 184 & 64 & 149 & 7,899 \\
Clips & 5,889 & 3,593 & 2,335 & 1,603 & 800 & 437 & 412 & 321 & 165 & 15,555 \\
\midrule
\textit{1. Clean capture} & 3,792 & 3,424 & 2,325 & 1,406 & 729 & 433 & 402 & 318 & 115 & 12,944 \\
\midrule
\multicolumn{11}{@{}l}{\textit{2. Arms can follow}} \\
\hspace*{6pt}SharpA & 1,835 & 1,828 & 1,183 & 378 & 581 & 404 & 93 & 237 & 115 & 6,654 \\
\hspace*{6pt}XHand Pro & 1,897 & 1,961 & 1,522 & 419 & 693 & 412 & 146 & 240 & 115 & 7,405 \\
\hspace*{6pt}Tesollo & 1,981 & 2,044 & 1,607 & 384 & 686 & 419 & 154 & 251 & 115 & 7,641 \\
\hspace*{6pt}Wuji 2 & 1,811 & 1,675 & 1,354 & 333 & 629 & 414 & 123 & 234 & 109 & 6,682 \\
\midrule
\multicolumn{11}{@{}l}{\textit{3. Reproduced in physics}} \\
\hspace*{6pt}SharpA & 1,375 & 1,434 & 1,018 & 303 & 211 & 384 & 83 & 222 & 101 & \textbf{5,131} \\
\hspace*{6pt}XHand Pro & 1,374 & 1,518 & 1,172 & 293 & 188 & 386 & 116 & 213 & 101 & \textbf{5,361} \\
\hspace*{6pt}Tesollo & 1,373 & 1,532 & 1,107 & 247 & 162 & 398 & 124 & 212 & 73 & \textbf{5,228} \\
\hspace*{6pt}Wuji 2 & 1,327 & 1,327 & 1,153 & 255 & 181 & 396 & 104 & 204 & 101 & \textbf{5,048} \\
\bottomrule
\end{tabular}

%% file: sections/floats/tab_supp_rollouts.tex
\begin{table}[t]
\centering
{\footnotesize\setlength{\tabcolsep}{4pt}
\begin{tabular}{@{}lrrrr@{}}
\toprule
 & & \multicolumn{3}{c}{Mean over rollouts} \\
\cmidrule(l){3-5}
Method & $K$ & ADD-AUC & SP-SR & MP-SR \\
\midrule
\multicolumn{5}{@{}l}{\textit{Floating hand}} \\
\hspace*{6pt}ManipTrans~\cite{maniptrans} & 1 & 0.51 & 39.4 & 15.6 \\
\hspace*{6pt}DexMachina~\cite{dexmachina} & 1 & 0.55 & 67.0 & 0.0 \\
\hspace*{6pt}SPIDER~\cite{spider} & 1 & 0.62 & 78.1 & 23.4 \\
\hspace*{6pt}CHORD~\cite{chord} & 1 & 0.76 & 77.7 & 61.3 \\
\hspace*{6pt}EgoEngine~\cite{egoengine} & 1 & 0.56 & 65.6 & 50.0 \\
\hspace*{6pt}FlashDexRetarget~\cite{flashdexretarget} & 64 & 0.77 & 79.1 & 55.6 \\
\hspace*{6pt}DexEngine (Ours) & 64 & 0.89 & 95.6 & 77.7 \\
\midrule
\multicolumn{5}{@{}l}{\textit{Arm-mounted hand}} \\
\hspace*{6pt}DexEngine (Ours) & 64 & 0.83 & 86.9 & 64.1 \\
\hspace*{6pt}DexEngine (Ours) & 32 & 0.91 & 98.8 & 83.4 \\
\hspace*{6pt}DexEngine (Ours) & 16 & 0.93 & 99.4 & 84.3 \\
\hspace*{6pt}DexEngine (Ours) & 4 & 0.95 & 99.1 & 91.0 \\
\hspace*{6pt}DexEngine (Ours) & 1 & 0.96 & 98.3 & 90.3 \\
\bottomrule
\end{tabular}}
\caption{\Cref{tab:same64} (SharpA hand) scored by the mean over the 64 rollouts of each clip, the expected score of one rollout. SPIDER and EgoEngine produce one trajectory per run, and their values equal those of \cref{tab:same64}.}
\label{tab:supp_rollouts}
\end{table}

%% file: sections/floats/tab_supp_posthoc.tex
\begin{table}[t]
\centering
{\footnotesize\setlength{\tabcolsep}{7pt}
\input{figs_tabs/tables/tab_supp_posthoc_body}}
\caption{Arm IK on the 64 clips of \cref{tab:same64} at the placements our search adopted, on the wrist targets of our reference and on those reached by our floating-hand policy and by SPIDER's rollouts.}
\label{tab:supp_posthoc}
\end{table}

%% file: figs_tabs/tables/tab_supp_posthoc_body.tex
\begin{tabular}{@{}lr@{}}
\toprule
Wrist targets of & Clips kept (\%) \\
\midrule
Our reference & 89 \\
Our floating-hand policy & 66 \\
SPIDER's rollouts & 50 \\
\bottomrule
\end{tabular}

%% file: sections/floats/tab_abl_place.tex
\begin{table}[t]
\centering
{\footnotesize\setlength{\tabcolsep}{7pt}
\input{figs_tabs/tables/tab_abl_place_body}}
\caption{Placement rules on the same 128 Wuji~2 clips. Kept: clips with a reference. IK: median candidates tested by the arm IK. Time: mean wall-clock seconds per clip. Random and every-candidate rows skip the screening.}
\label{tab:abl_place}
\end{table}

%% file: figs_tabs/tables/tab_abl_place_body.tex
\begin{tabular}{@{}lrrr@{}}
\toprule
Placements & Kept (\%) & IK & Time (s) \\
\midrule
Default position & 1.6 & 1 & 6.7 \\
12 random & 8.6 & 12 & 7.5 \\
Top screened & 21.9 & 1 & 10.3 \\
\midrule
Random, up to 1,024 tested & 34.4 & 1,024 & 24.1 \\
Random, up to 1,200\,s & 45.3 & 18,793 & 204 \\
Every candidate & 45.3 & 12,808 & 222 \\
Ours & 43.8 & 784 & 68.7 \\
\bottomrule
\end{tabular}

%% file: sections/floats/tab_supp_ablation.tex
\begin{table*}[t]
\centering
{\footnotesize\setlength{\tabcolsep}{4pt}
\input{figs_tabs/tables/tab_supp_ablation_body}}
\caption{Changes of the RL-design ablation (SharpA hand), with the setup and scoring of \cref{tab:abl_rl}. Best: best of the 64 rollouts of a clip for ADD-AUC, any of them for SP-SR and MP-SR. Mean: over the rollouts. Iter.: training iterations.}
\label{tab:supp_ablation}
\end{table*}

%% file: figs_tabs/tables/tab_supp_ablation_body.tex
\begin{tabular}{@{}lrrrrrrr@{}}
\toprule
 & \multicolumn{2}{c}{ADD-AUC} & \multicolumn{2}{c}{SP-SR} & \multicolumn{2}{c}{MP-SR} & \\
\cmidrule(lr){2-3}\cmidrule(lr){4-5}\cmidrule(lr){6-7}
Change from the full recipe & best & mean & best & mean & best & mean & Iter. \\
\midrule
None (full recipe) & 0.900 & 0.839 & 85.9 & 69.6 & 81.2 & 67.5 & 1,600 \\
\midrule
\multicolumn{8}{@{}l}{\textit{Random starts}} \\
No random starts (first frame only) & 0.888 & 0.827 & 82.8 & 65.2 & 79.7 & 63.8 & 1,600 \\
Arm re-solved by IK at random starts & 0.832 & 0.748 & 75.0 & 53.5 & 71.9 & 50.5 & 1,600 \\
\multicolumn{8}{@{}l}{\textit{Arm action and tracking}} \\
Residuals on the arm joints instead of the wrist target & 0.856 & 0.795 & 76.6 & 61.3 & 73.4 & 57.9 & 1,600 \\
Residuals on the arm joints, arm re-solved by IK at random starts & 0.544 & 0.500 & 25.0 & 18.5 & 20.3 & 16.5 & 1,600 \\
No low-pass filter before IK & 0.885 & 0.811 & 85.9 & 62.9 & 82.8 & 57.8 & 1,600 \\
\multicolumn{8}{@{}l}{\textit{Termination and reward terms}} \\
Strict termination at the current frame, no disarmed period & 0.898 & 0.834 & 84.4 & 66.7 & 81.2 & 63.3 & 1,600 \\
No contact terms & 0.837 & 0.780 & 76.6 & 64.9 & 70.3 & 62.6 & 1,600 \\
No fingertip tolerance & 0.883 & 0.831 & 79.7 & 68.0 & 78.1 & 63.3 & 1,600 \\
No action-rate penalties & 0.903 & 0.823 & 87.5 & 68.3 & 82.8 & 65.1 & 1,600 \\
\multicolumn{8}{@{}l}{\textit{Schedules and curricula}} \\
Final termination thresholds from the start & 0.897 & 0.841 & 82.8 & 69.2 & 79.7 & 64.3 & 1,600 \\
No object hold after a reset & 0.882 & 0.804 & 79.7 & 62.8 & 76.6 & 59.7 & 1,600 \\
No friction decay & 0.886 & 0.830 & 82.8 & 67.4 & 81.2 & 64.5 & 1,600 \\
\midrule
\multicolumn{8}{@{}l}{\textit{Further changes}} \\
No domain randomization & 0.900 & 0.850 & 84.4 & 71.0 & 81.2 & 68.1 & 1,600 \\
Objects at rest at random starts & 0.895 & 0.837 & 89.1 & 70.5 & 82.8 & 65.2 & 1,600 \\
KL threshold 0.02 instead of 0.04 & 0.756 & 0.659 & 43.8 & 27.3 & 42.2 & 24.6 & 1,600 \\
2,048 parallel environments instead of 4,096 & 0.855 & 0.778 & 73.4 & 55.7 & 70.3 & 52.2 & 1,600 \\
Actor and critic of width 128 and four layers & 0.758 & 0.629 & 60.9 & 35.8 & 53.1 & 32.8 & 1,600 \\
Actor and critic of width 96 & 0.886 & 0.824 & 81.2 & 64.0 & 76.6 & 58.9 & 1,600 \\
\bottomrule
\end{tabular}

%% file: sections/floats/tab_supp_robust.tex
\begin{table}[t]
\centering
\fitwidth{\footnotesize\setlength{\tabcolsep}{4pt}\input{figs_tabs/tables/tab_supp_robust_body}}
\caption{The $K=64$ arm-mounted SharpA policies of \cref{tab:same64}, re-evaluated without retraining under changed physics (\cref{sec:supp_abl_rl}). Row 1 repeats \cref{tab:same64}. In the friction row, object--table friction falls to 0.6 of its value.}
\label{tab:supp_robust}
\end{table}

%% file: figs_tabs/tables/tab_supp_robust_body.tex
\begin{tabular}{@{}lrrr@{}}
\toprule
Evaluated with & ADD-AUC & SP-SR & MP-SR \\
\midrule
Training conditions & 0.897 & 98.4 & 84.4 \\
Manipulated-object mass halved & 0.879 & 98.4 & 84.4 \\
Manipulated-object mass doubled & 0.890 & 96.9 & 84.4 \\
Hand--object friction halved & 0.861 & 95.3 & 73.4 \\
\bottomrule
\end{tabular}

%% file: sections/floats/tab_supp_e5_long.tex
\begin{table}[t]
\centering
{\footnotesize\setlength{\tabcolsep}{0pt}
\input{figs_tabs/tables/tab_supp_e5_long_body}}
\caption{GR00T success with Wuji~2 hands in simulation (\%) at two fine-tuning lengths, on the scenes of \cref{tab:main}. About 150 passes: 199,000, 187,000 and 140,000 steps for P\&P, Pour and Sweep, as on kinematic replay. Bold: best.}
\label{tab:supp_e5_long}
\end{table}

%% file: figs_tabs/tables/tab_supp_e5_long_body.tex
\begin{tabular*}{\linewidth}{@{\extracolsep{\fill}}l*{3}{r@{\extracolsep{0pt}\hspace{7pt}}r@{\extracolsep{\fill}}}@{\hspace{5pt}}r@{\extracolsep{0pt}\hspace{7pt}}r@{}}
\toprule
 & \multicolumn{2}{c}{P\&P} & \multicolumn{2}{c}{Pour} & \multicolumn{2}{c}{Sweep} & \multicolumn{2}{c}{Mean} \\
\cmidrule(lr){2-3} \cmidrule(lr){4-5} \cmidrule(lr){6-7} \cmidrule(l){8-9}
Policy & L0 & L1 & L0 & L1 & L0 & L1 & L0 & L1 \\
\midrule
\multicolumn{9}{@{}l}{\textit{Kinematic replay data, about 150 passes}} \\
\hspace*{6pt}GR00T~\cite{gr00t} & 12 & 8 & 3 & 4 & 3 & 3 & 6 & 5 \\
\midrule
\multicolumn{9}{@{}l}{\textit{DexEngine data (Ours), 40,000 steps}} \\
\hspace*{6pt}GR00T & 50 & 45 & 74 & 52 & 70 & 54 & 65 & 50 \\
\hspace*{6pt}GR00T + mid & 72 & 48 & 75 & \textbf{61} & 81 & \textbf{62} & 76 & \textbf{57} \\
\midrule
\multicolumn{9}{@{}l}{\textit{DexEngine data (Ours), about 150 passes}} \\
\hspace*{6pt}GR00T & \textbf{78} & 45 & 80 & 47 & \textbf{88} & 58 & \textbf{82} & 50 \\
\hspace*{6pt}GR00T + mid & 74 & \textbf{54} & \textbf{82} & 53 & 86 & 57 & 81 & 55 \\
\bottomrule
\end{tabular*}

%% file: sections/floats/fig_supp_realaug.tex
\begin{figure*}[t]
  \centering
  \includegraphics[width=\linewidth]{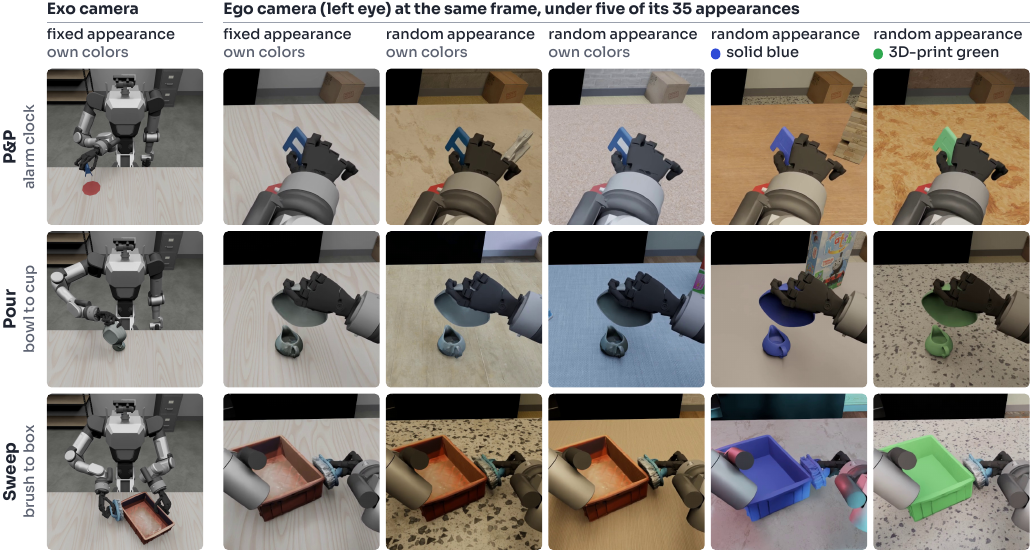}
  \caption{Visual augmentation of the real-robot fine-tuning data (Wuji~2 hands).
  Rows: the P\&P, Pour and Sweep clips of \cref{tab:real}, one placement variant each.
  Left: the exo camera under the fixed appearance.
  Right: the left-eye ego camera at the same frame under five of its 35 appearances, which randomize the materials, lighting and distractor objects and change the colors of the manipulated objects, including the green of our 3D prints.}
  \label{fig:supp_realaug}
\end{figure*}

%% file: sections/floats/tab_supp_drops.tex
\begin{table}[t]
\centering
{\footnotesize\setlength{\tabcolsep}{4pt}
\begin{tabular}{@{}lrrrr@{}}
\toprule
Clips & SharpA & XHand Pro & Tesollo & Wuji~2 \\
\midrule
Tested & 12,929 & 12,950 & 12,943 & 12,951 \\
Kept & 6,654 & 7,405 & 7,641 & 6,682 \\
\midrule
Screened candidates ran out & 1,765 & 1,998 & 2,129 & 2,247 \\
Limit on candidates tested & 4,331 & 3,325 & 2,958 & 3,799 \\
Time limit & 25 & 4 & 0 & 4 \\
Failed a check after the search & 154 & 218 & 215 & 219 \\
\bottomrule
\end{tabular}}
\caption{End of the placement search for every clip it tested, per robot hand (limits in \cref{sec:supp_order}). The last row counts clips with passing candidates that fail a later check.}
\label{tab:supp_drops}
\end{table}

%% file: sections/floats/tab_supp_selfpen.tex
\begin{table}[t]
\centering
{\footnotesize\setlength{\tabcolsep}{3pt}
\input{figs_tabs/tables/tab_supp_selfpen_body}}
\caption{Trajectories with a penetration deeper than 2\,cm, in \% of each hand's trajectories, per pair (\cref{sec:supp_recording}). Hand--arm: a hand, including the last arm link, and a link of the other arm. Excluded: trajectories with any such pair, counted once.}
\label{tab:supp_selfpen}
\end{table}

%% file: figs_tabs/tables/tab_supp_selfpen_body.tex
\begin{tabular}{@{}lrrrrr@{}}
\toprule
Hand & Arm--arm & Arm--torso & Hand--arm & Hand--torso & Excluded \\
\midrule
Wuji~2 & 0.3 & 0.3 & 0.5 & $<$0.1 & 1.0 \\
SharpA & 0.6 & 2.5 & 0.6 & 0.2 & 3.3 \\
XHand~Pro & 0.2 & 3.1 & 0.7 & 0.2 & 4.0 \\
Tesollo & 2.4 & 2.1 & 1.0 & 0.0 & 5.2 \\
\bottomrule
\end{tabular}

%% file: sections/floats/tab_supp_rl_more.tex
\begin{table*}[p]
\centering
{\footnotesize\setlength{\tabcolsep}{4pt}
\input{figs_tabs/tables/tab_supp_rl_more_body}}
\caption{RL recipe built up on the arm-mounted SharpA hand (a), changed on the floating hand (b), with further single changes (c) and trained again (d), on the 64 clips of \cref{tab:same64} with $K=64$, 1,600 iterations and the scoring of \cref{tab:abl_rl}. Best and mean as in \cref{tab:supp_ablation}. The rows of (b) compare only with each other.}
\label{tab:supp_rl_more}
\end{table*}

%% file: figs_tabs/tables/tab_supp_rl_more_body.tex
\begin{tabular}{@{}>{\raggedright\arraybackslash}p{10.6cm}rrrrrr@{}}
\toprule
 & \multicolumn{2}{c}{ADD-AUC} & \multicolumn{2}{c}{SP-SR} & \multicolumn{2}{c}{MP-SR} \\
\cmidrule(lr){2-3}\cmidrule(lr){4-5}\cmidrule(l){6-7}
Training & best & mean & best & mean & best & mean \\
\midrule
\multicolumn{7}{@{}l}{\textit{(a) Arm-mounted hand, each row added to the row above}} \\
Plain tracking & 0.748 & 0.688 & 65.6 & 49.8 & 60.9 & 47.0 \\
+ gravity compensation, velocity lead, termination window, curriculum, value per hand, penalties, randomization & 0.590 & 0.538 & 35.9 & 23.2 & 29.7 & 21.3 \\
+ random starts, arm re-solved by IK & 0.629 & 0.558 & 45.3 & 29.3 & 43.8 & 28.1 \\
Random starts with the arm reset to our reference instead & 0.789 & 0.713 & 65.6 & 53.3 & 62.5 & 48.7 \\
+ contact terms & 0.865 & 0.791 & 84.4 & 63.8 & 81.2 & 59.4 \\
+ object hold after a reset, friction decay & 0.881 & 0.827 & 79.7 & 64.0 & 75.0 & 58.7 \\
+ fingertip tolerance, low-pass filter before IK (full recipe) & 0.900 & 0.839 & 85.9 & 69.6 & 81.2 & 67.5 \\
\midrule
\multicolumn{7}{@{}l}{\textit{(b) Floating hand, one change from its full recipe}} \\
None (full recipe) & 0.891 & 0.847 & 82.8 & 71.5 & 78.1 & 67.9 \\
No random starts (first frame only) & 0.891 & 0.830 & 81.2 & 64.3 & 81.2 & 60.6 \\
No contact terms & 0.853 & 0.820 & 79.7 & 73.9 & 75.0 & 68.9 \\
Strict termination at the current frame, no disarmed period & 0.894 & 0.852 & 82.8 & 72.3 & 79.7 & 68.9 \\
Summed reward with a shared value head & 0.894 & 0.845 & 89.1 & 71.9 & 84.4 & 66.1 \\
Plain tracking & 0.835 & 0.777 & 75.0 & 66.3 & 71.9 & 62.2 \\
\midrule
\multicolumn{7}{@{}l}{\textit{(c) Arm-mounted hand, further single changes}} \\
Wrist position residual bounded at 28\,cm instead of 7\,cm & 0.904 & 0.833 & 84.4 & 66.2 & 78.1 & 62.4 \\
Wrist position residual bounded at 112\,cm instead of 7\,cm & 0.909 & 0.850 & 90.6 & 79.9 & 85.9 & 76.1 \\
Absolute wrist and finger targets instead of residuals & 0.910 & 0.839 & 93.8 & 79.2 & 92.2 & 72.4 \\
A separate critic per hand & 0.899 & 0.836 & 90.6 & 66.9 & 85.9 & 63.4 \\
Summed reward with a shared value head & 0.896 & 0.843 & 90.6 & 76.0 & 84.4 & 71.4 \\
Penalties on joint power added & 0.902 & 0.837 & 90.6 & 71.4 & 87.5 & 67.3 \\
No per-element wrist-residual free band & 0.900 & 0.851 & 85.9 & 71.4 & 84.4 & 67.8 \\
No look-ahead in the wrist target or the termination window & 0.921 & 0.855 & 90.6 & 71.1 & 89.1 & 66.8 \\
No arm gravity compensation or velocity lead & 0.907 & 0.852 & 87.5 & 70.6 & 84.4 & 66.8 \\
Arm re-solved by IK at random starts, no look-ahead, gravity compensation, velocity lead or low-pass filter & 0.863 & 0.794 & 82.8 & 62.7 & 81.2 & 58.6 \\
No gravity ramp on the objects & 0.916 & 0.857 & 85.9 & 72.3 & 81.2 & 67.6 \\
8 mini-batches instead of 16 & 0.926 & 0.868 & 92.2 & 75.3 & 89.1 & 71.8 \\
Reference arm from one sequential IK run, first frame from our solution & 0.907 & 0.830 & 89.1 & 70.6 & 85.9 & 65.7 \\
Reference arm from one sequential IK run approaching from the rest pose, our timing kept & 0.885 & 0.829 & 81.2 & 61.1 & 76.6 & 55.4 \\
\midrule
\multicolumn{7}{@{}l}{\textit{(d) Arm-mounted hand, the same training repeated}} \\
Full recipe & 0.900 & 0.839 & 85.9 & 69.6 & 81.2 & 67.5 \\
 & 0.896 & 0.836 & 84.4 & 66.6 & 81.2 & 63.3 \\
 & 0.900 & 0.837 & 89.1 & 68.9 & 82.8 & 65.4 \\
No random starts (first frame only) & 0.888 & 0.827 & 82.8 & 65.2 & 79.7 & 63.8 \\
 & 0.892 & 0.834 & 79.7 & 63.7 & 75.0 & 59.1 \\
 & 0.883 & 0.814 & 79.7 & 58.0 & 75.0 & 54.2 \\
Strict termination at the current frame, no disarmed period & 0.898 & 0.834 & 84.4 & 66.7 & 81.2 & 63.3 \\
 & 0.898 & 0.843 & 84.4 & 69.8 & 82.8 & 65.4 \\
No contact terms & 0.837 & 0.780 & 76.6 & 64.9 & 70.3 & 62.6 \\
 & 0.838 & 0.780 & 78.1 & 62.7 & 75.0 & 57.0 \\
No fingertip tolerance & 0.883 & 0.831 & 79.7 & 68.0 & 78.1 & 63.3 \\
 & 0.888 & 0.819 & 89.1 & 66.7 & 85.9 & 63.1 \\
No gravity ramp on the objects & 0.916 & 0.857 & 85.9 & 72.3 & 81.2 & 67.6 \\
 & 0.897 & 0.844 & 85.9 & 69.7 & 81.2 & 67.2 \\
No look-ahead in the wrist target or the termination window & 0.921 & 0.855 & 90.6 & 71.1 & 89.1 & 66.8 \\
 & 0.911 & 0.856 & 89.1 & 70.9 & 84.4 & 66.5 \\
Absolute wrist and finger targets instead of residuals & 0.899 & 0.841 & 90.6 & 80.3 & 87.5 & 76.2 \\
 & 0.910 & 0.839 & 93.8 & 79.2 & 92.2 & 72.4 \\
\bottomrule
\end{tabular}

%% file: sections/S6_supp_gallery.tex
\section{Data samples}
\label{sec:supp_gallery}
\Cref{fig:supp_gallery_episode} shows every stream of one recorded episode, in the rows of \cref{tab:spec}.
\Cref{fig:supp_gallery_aug} shows what pose and visual augmentation change in these streams.
\Cref{fig:supp_gallery_datasets,fig:supp_gallery_hands} show episodes from the nine source datasets and on the four robot hands.
\input{sections/floats/fig_supp_gallery_episode}
\input{sections/floats/fig_supp_gallery_aug}
\input{sections/floats/fig_supp_gallery_datasets}
\input{sections/floats/fig_supp_gallery_hands}
\clearpage
\makeatletter\global\@colroom\@colht\global\vsize\@colht\makeatother

%% file: sections/floats/fig_supp_gallery_episode.tex
\begin{figure*}[p]
  \centering
  \includegraphics[width=\linewidth]{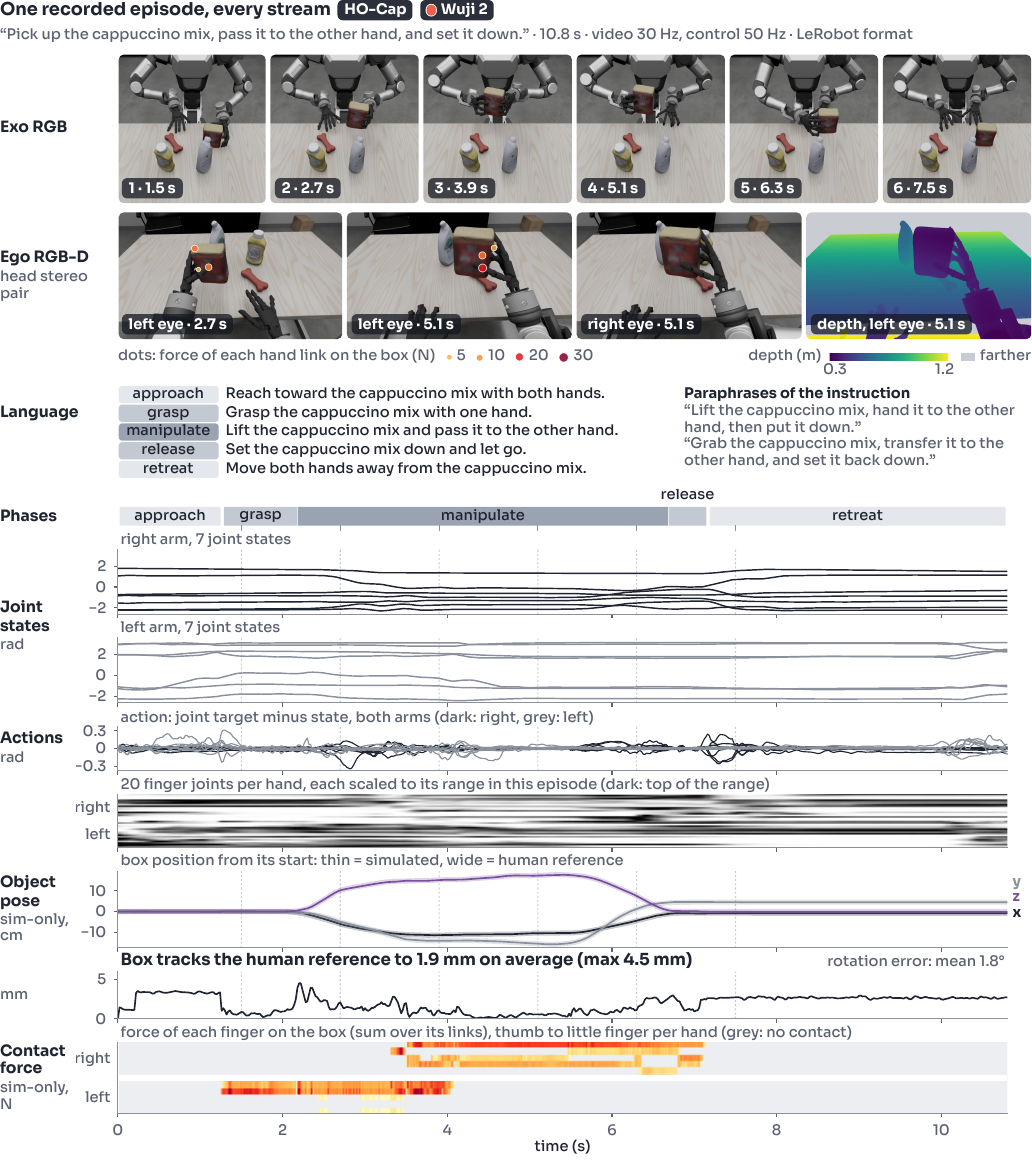}
  \caption{One recorded episode (HO-Cap clip, Wuji~2 hands) with all its streams, in the rows of \cref{tab:spec}.
  The numbered exo frames mark their times on the plots.
  Dots on the left-eye frames and the bottom strip give the force of the hands on the box, which passes from the left hand to the right.
  Sim-only marks the privileged streams of \cref{tab:spec}, which only the simulator provides.}
  \label{fig:supp_gallery_episode}
\end{figure*}

%% file: sections/floats/fig_supp_gallery_aug.tex
\begin{figure*}[p]
  \centering
  \includegraphics[width=\linewidth]{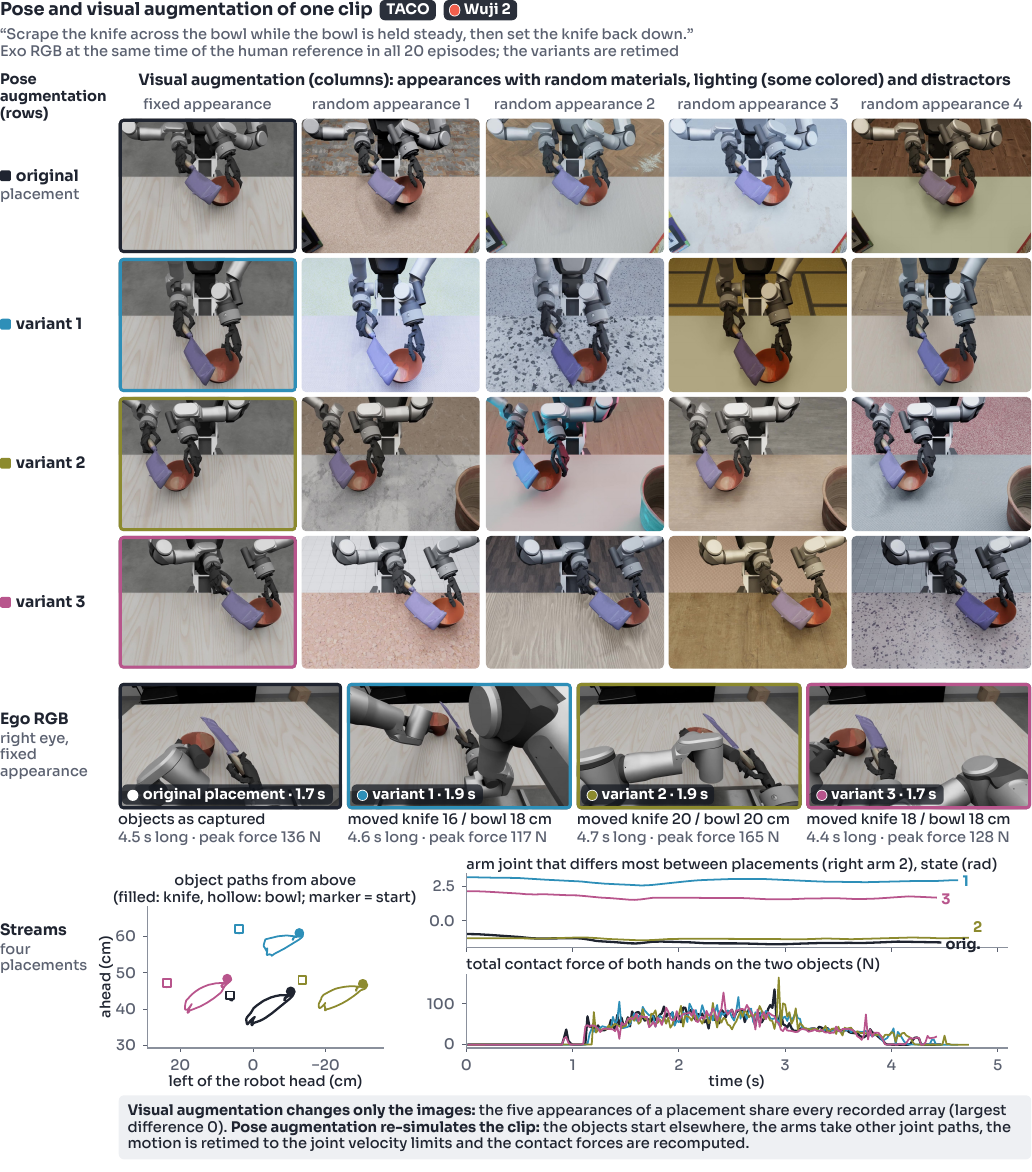}
  \caption{Pose and visual augmentation of one two-handed TACO clip (Wuji~2 hands): the right hand scrapes a knife across the bowl that the left hand holds.
  Rows: the original placement and three pose-augmentation variants, each simulated again.
  Columns: the fixed and the four random appearances of visual augmentation, which change only the images.}
  \label{fig:supp_gallery_aug}
\end{figure*}

%% file: sections/floats/fig_supp_gallery_datasets.tex
\begin{figure*}[p]
  \centering
  \includegraphics[width=\linewidth]{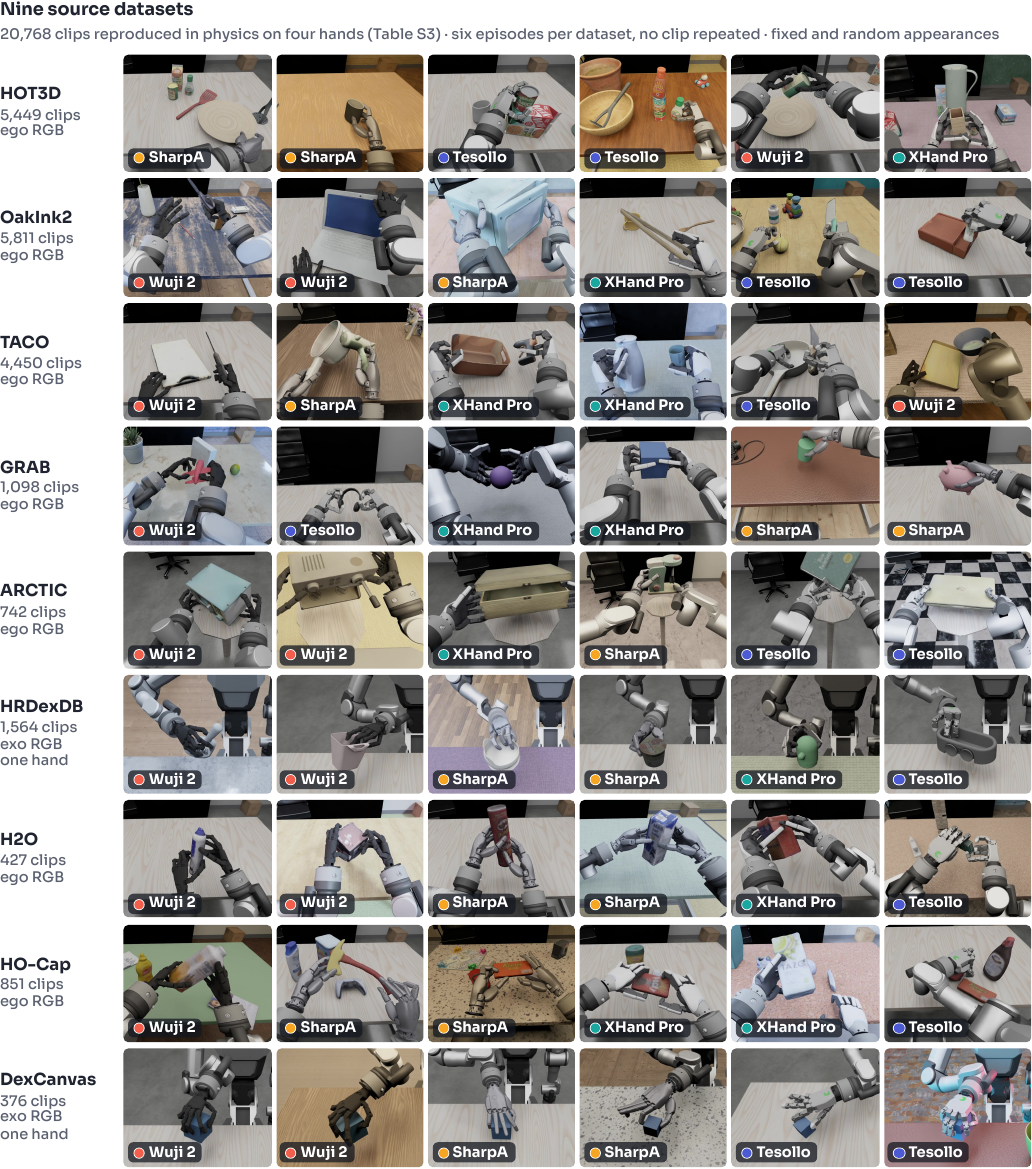}
  \caption{Recorded episodes from the nine source datasets on the four robot hands, half of them under a random appearance of visual augmentation.
  Ego view, but exo view for the one-hand clips of HRDexDB and DexCanvas, where the arm hides the hand from the head camera.}
  \label{fig:supp_gallery_datasets}
\end{figure*}

%% file: sections/floats/fig_supp_gallery_hands.tex
\begin{figure*}[p]
  \centering
  \includegraphics[width=\linewidth]{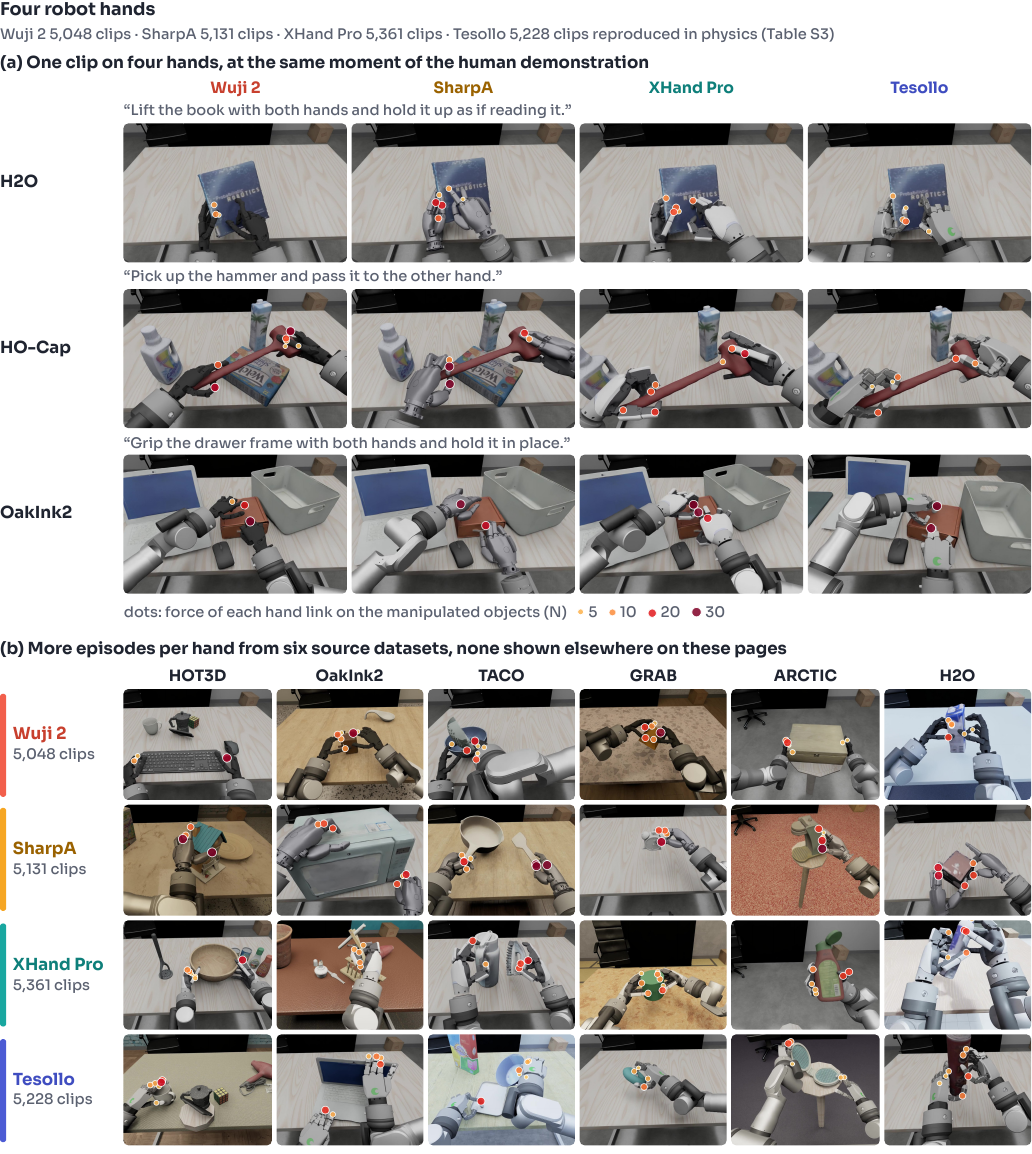}
  \caption{The four robot hands.
  (a)~Three two-handed clips, each on the four hands at the same moment of the human demonstration.
  (b)~Episodes from six source datasets per hand, none of them shown elsewhere in this section.
  Dots: force of each hand link on the manipulated objects, on the scale of \cref{fig:supp_gallery_episode}.}
  \label{fig:supp_gallery_hands}
\end{figure*}